\documentclass{article}
\PassOptionsToPackage{numbers, compress}{natbib}
\usepackage{neurips_2024}

\usepackage[utf8]{inputenc} 
\usepackage[T1]{fontenc}    
\usepackage{microtype}      
\usepackage{url}            
\usepackage{graphicx}
\usepackage{subfigure}
\usepackage{booktabs}       
\usepackage{multirow}
\usepackage{makecell}

\usepackage{floatrow}

\usepackage{researchpack}
\usepackage{hyperref}       
\usepackage[capitalize, nameinlink]{cleveref}
\usepackage{pifont}
\usepackage{enumitem}
\usepackage{amsfonts}       
\usepackage{nicefrac}       
\usepackage[dvipsnames]{xcolor}
\usepackage{colortbl} 
\usepackage{caption}
\usepackage{subcaption}
\usepackage{tablefootnote}
\usepackage{soul} 
\usepackage{wrapfig} 
\usepackage{tikz} 
\usepackage{pifont} 

\hypersetup{
    colorlinks=true,
    linkcolor=MidnightBlue,
    citecolor=MidnightBlue,
    urlcolor=MidnightBlue
}

\usepackage{thmtools, thm-restate}
\declaretheorem[style=plain]{definition, theorem, proposition, lemma}

\theoremstyle{plain}
\newtheorem{defn}{Definition}

\newtheorem{example}{Example}

\newcommand{\SA}[1]{\textcolor{orange}{\textbf{[SA: #1]}}\xspace}

\newcommand{\GX}{\ensuremath{G_A}\xspace}
\newcommand{\RX}{\ensuremath{R_A}\xspace}
\newcommand{\CX}{\ensuremath{C_A}\xspace}
\newcommand{\GXInv}{\ensuremath{\RX^{*}}\xspace}

\newcommand{\expectedGy}[1]{\bbE_{(\GX, y) \sim #1}}
\newcommand{\expectedCondGy}[2]{\bbE_{(\GX, y) \sim #1(\GX, Y \mid #2)}}
\newcommand{\expectedCondInvGy}[1]{\bbE_{(\GX, y) \sim #1(\cdot \mid \GXInv)}}

\newcommand{\upd}{\ensuremath{\mathsf{update}}\xspace}
\newcommand{\aggr}{\ensuremath{\mathsf{aggr}}\xspace}
\newcommand{\MONO}{\ensuremath{{p_\theta}}\xspace}
\newcommand{\DET}{\ensuremath{f}\xspace} 
\newcommand{\CLF}{\ensuremath{g}\xspace} 

\newcommand{\supp}{\ensuremath{\mathrm{supp}}\xspace}
\newcommand{\KL}{\ensuremath{\mathsf{KL}}\xspace}

\newcommand{\PR}{\ensuremath{p_{R}}\xspace}
\newcommand{\PC}{\ensuremath{p_{C}}\xspace}

\newcommand{\NEC}{\ensuremath{\mathsf{Nec}}\xspace}
\newcommand{\SUF}{\ensuremath{\mathsf{Suf}}\xspace}
\newcommand{\FAITH}{\ensuremath{\mathsf{Faith}}\xspace}

\newcommand{\UNF}{\ensuremath{\mathsf{Unf}}\xspace}
\newcommand{\FIDP}{\ensuremath{\mathsf{Fid}\scalebox{0.9}{+}}\xspace}
\newcommand{\FIDM}{\ensuremath{\mathsf{Fid}\scalebox{0.9}{-}}\xspace}
\newcommand{\RFIDP}{\ensuremath{\mathsf{RFid}\scalebox{0.9}{+}}\xspace}
\newcommand{\RFIDM}{\ensuremath{\mathsf{RFid}\scalebox{0.9}{-}}\xspace}
\newcommand{\PROBSUFF}{\ensuremath{\mathsf{PS}}\xspace}
\newcommand{\PROBNEC}{\ensuremath{\mathsf{PN}}\xspace}
\newcommand{\ADVROB}{\ensuremath{\mathsf{AdvRob}}\xspace}

\newcommand{\ACC}{\ensuremath{\mathsf{Acc}}\xspace}
\newcommand{\LAMBDATOPO}{\ensuremath{\lambda_{\text{topo}}}\xspace}
\newcommand{\LAMBDAFEAT}{\ensuremath{\lambda_{\text{feat}}}\xspace}
\newcommand{\LAMBDASUFF}{\ensuremath{\lambda_{\text{suff}}}\xspace}

\newcommand{\DIGNN}{DI-GNN\xspace}
\newcommand{\DIGNNs}{DI-GNNs\xspace}

\newcommand{\SEGNNs}{SE-GNNs\xspace}

\newcommand{\RAGE}{\texttt{RAGE}\xspace}
\newcommand{\GISST}{\texttt{GISST}\xspace}

\newcommand{\GIN}{\texttt{GIN}\xspace}

\newcommand{\GSAT}{\texttt{GSAT}\xspace}

\newcommand{\LECI}{\texttt{LECI}\xspace}
\newcommand{\CIGA}{\texttt{CIGA}\xspace}

\newcommand{\BBBP}{\texttt{BBBP}\xspace}

\newcommand{\BAMULTISHAPES}{\texttt{BaMS}\xspace}
\newcommand{\MotifBasis}{\texttt{Motif-Basis}\xspace}
\newcommand{\MotifTwoBasis}{\texttt{Motif2-Basis}\xspace}
\newcommand{\MotifTwoBasisS}{\texttt{Motif2}\xspace}
\newcommand{\MotifSize}{\texttt{Motif-Size}\xspace}
\newcommand{\SSTTwo}{\texttt{SST2-Length}\xspace}
\newcommand{\SSTTwoS}{\texttt{SST2}\xspace}

\newcommand{\LBAPcore}{\texttt{LBAPcore-Assay}\xspace}
\newcommand{\LBAPS}{\texttt{LBAPcore}\xspace}
\newcommand{\CMNIST}{\texttt{CMNIST-Color}\xspace}
\newcommand{\CMNISTS}{\texttt{CMNIST}\xspace}

\newcommand{\budget}{b\xspace}
\newcommand{\budgetMC}{\ensuremath{q_1}\xspace}
\newcommand{\budgetSamples}{\ensuremath{q_2}\xspace}

\newcommand{\cmark}{\textcolor{ForestGreen}{\ding{51}}\xspace}
\newcommand{\xmark}{\textcolor{BrickRed}{\ding{55}}\xspace}

\newcommand{\nentry}[2]{#1 \text{\tiny{$\pm$ #2}}}

\newcommand{\rentry}[2]{\textcolor{BrickRed}{#1 \text{\tiny{$\pm$ #2}}}}
\newcommand{\gentry}[2]{\textcolor{ForestGreen}{#1 \text{\tiny{$\pm$ #2}}}}
\newcommand{\correction}[2]{\textcolor{red}{\st{#1} #2}}

\author{%
  David S.~Hippocampus\thanks{Use footnote for providing further information
    about author (webpage, alternative address)---\emph{not} for acknowledging
    funding agencies.} \\
  Department of Computer Science\\
  Cranberry-Lemon University\\
  Pittsburgh, PA 15213 \\
  \texttt{hippo@cs.cranberry-lemon.edu} \\
}

\begin{document}
\maketitle

\begin{abstract}
    As Graph Neural Networks (GNNs) become more pervasive, it becomes paramount to build reliable tools for explaining their predictions.
    A core desideratum is that explanations are \textit{faithful}, \ie that they portray an accurate picture of the GNN's reasoning process.
    However, a number of different faithfulness metrics exist, begging the question of what is faithfulness exactly and how to achieve it.
    We make three key contributions.
    We begin by showing that \textit{existing metrics are not interchangeable} -- \ie explanations attaining high faithfulness according to one metric may be unfaithful according to others -- and can \textit{systematically ignore important properties of explanations}.
    We proceed to show that, surprisingly, \textit{optimizing for faithfulness is not always a sensible design goal}.  Specifically, we prove that for injective regular GNN architectures, perfectly faithful explanations are completely uninformative.
    This does not apply to modular GNNs, such as self-explainable and domain-invariant architectures, prompting us to study the relationship between architectural choices and faithfulness.
    Finally, we show that \textit{faithfulness is tightly linked to out-of-distribution generalization}, in that simply ensuring that a GNN can correctly recognize the domain-invariant subgraph, as prescribed by the literature, does not guarantee that it is invariant unless this subgraph is also faithful.
    All our code can be found in the supplementary material.
\end{abstract}

\section{Introduction}

The increasing popularity of GNNs \citep{scarselli2008graph, kipf2016semi, velivckovic2018graph} for even high-stakes tasks \citep{agarwal2023evaluating} has prompted the development of tools for explaining their decisions.
Regular GNNs are opaque and explanations must be extracted in a \textit{post-hoc} fashion using specialized algorithms \citep{longa2022explaining}, while self-explainable GNNs employ a modular setup to natively output explanations for their own predictions \citep{kakkad2023survey}.

A key metric of explanation quality is \textit{\textbf{faithfulness}} \citep{agarwal2023evaluating, pope2019explainability, yuan2022explainability, amara2022graphframex, Zhao2022consistent, amara2023ginxeval, zheng2023robust, Juntao2022CF2}.  Intuitively, an explanation is \textit{faithful insofar as it highlights all and only those elements -- edges and/or node features -- of the input graph that are truly relevant for the prediction}.
Faithful explanations portray an accurate picture of the GNN's reasoning process, and thus supports understanding, trust modulation, and debugging \citep{teso2023leveraging}.

Existing faithfulness metrics assess the stability of the model's output to perturbations of the input graph -- \eg to ensure that modifying those elements that are marked as irrelevant by the explanation has in fact no effect -- yet differ in many non-trivial details.
For instance, some metrics perturb the input by zeroing-out all irrelevant node features \citep{agarwal2023evaluating} while others delete a subset of edges instead \citep{zheng2023robust};  cf. \cref{sec:faithfulness} for an overview.
Unfortunately, the literature provides little guidance on what metric should be used and on how to tune GNN architectures for faithfulness.
We aim to fill this gap.

\textbf{Not all metrics are the same.}  (\cref{sec:faithfulness})  We begin by parameterizing existing metrics along two dimensions:  how \textit{stability} is measured and what \textit{perturbations} are allowed.
We show that different parameter choices yield rather different metrics, in the sense that \textit{\textbf{explanations that are faithful for one metric are not faithful according to others}}.
These differences are far reaching.  In fact, we analyze popular metrics in detail and uncover systematic issues with some of them.
Specifically, we show they can be \textit{\textbf{insensitive to the number of irrelevant elements captured by the explanation}} and how to avoid this issue by tuning the parameters.
It follows that faithfulness measurements cannot be interpreted unless these parameters are known.
This is relevant in high-stakes applications, like loan approval, where explanation providers could manipulate (but withhold) the parameters to mislead end-users into overestimating the faithfulness of explanations \citep{bordt2022post}.

\textbf{Is faithfulness always worth optimizing for?}  (\cref{sec:perfect-faith})  Then, we study to what extent optimizing for faithfulness is a sensible design goal.
We show that for \textit{\textbf{injective GNNs, explanations achieving perfect faithfulness are not informative}}, and identify a natural trade-off between expressiveness of the model and usefulness of faithful explanations.
At the same time, we show that for self-explainable GNNs and domain invariant GNNs, which employ a modular architecture, strict faithful explanations \textit{can} be informative, and evaluate how architectural design choices can impact faithfulness, highlighting a trade-off between strict faithfulness and learnability. \SA{Questa è ancora la prospettiva "vecchia": ha senso pensare a "highlighting an implicit inductive bias favoring leakage of information between complement and  explanation"?}

\textbf{Faithfulness is key to OOD generalization.} (\cref{sec:sufficiency-invariance}) Finally, we highlight the central but so far neglected role of faithfulness in \textit{domain invariance}.
Specifically, prior work \cite{cai2023learning, chen2023rethinking, gui2023joint, jiang2023survey} tackles domain invariance by constructing GNNs that isolate the domain invariant portion of the input and use it for prediction.
We show that \textit{\textbf{\DIGNNs cannot be domain invariant unless the subgraphs they identify -- in addition to being invariant -- are also faithful}}, highlighting a key limitation in current design and evaluation strategies for domain-invariant GNNs, which neglect faithfulness altogether.

\section{Graph Neural Networks and Faithfulness}
\label{sec:background}

Throughout, we indicate graphs as $G = (V, E)$ and annotated graphs as $\GX = (G, X)$, where $\vx_u \in \bbR^d$ are per-node features.
We denote multisets as $\{\!\{ \ldots \}\!\}$, the $k$-hop neighborhood of $u$ as $N_k(u)$, and shorten $N_1(u)$ to $N(u)$, $|G| = |V|$ to $n$, and $\norm{G} = |E|$ to $m$.

\textbf{Graph Neural Networks} (GNN) \citep{scarselli2008graph} are discriminative classifiers that, given an input graph $\GX$, define a conditional distribution $p_\theta(\cdot \mid \GX)$ over candidate labels.
In graph classification, the label $y \in \{1, \ldots, c\}$ applies to the whole graph, while in node classification it is a vector $\vy \in \{1, \ldots, c\}^n$ with one element per node.
Inference in GNNs is \textit{opaque}, in that it relies on message passing of embedding vectors along the graph's topology.
Usually, this amounts to recursively applying an update-aggregate operation of the form:
\[
    \vh^\ell_u = \upd(\vh^{\ell - 1}_u, \aggr(\{\!\{ \vh^{\ell - 1}_v \ : \ v \in N(u) \}\!\}))
    \label{eq:message-passing}
\]
to all nodes $u$, from the bottom to the top layer.  Here, $\vh^0_u = \vx_u$ are the node features, $\vh^\ell_u$ the node embeddings at the $\ell$-th layer, $\upd$ a learnable non-linear function, and $\aggr$ a permutation-invariant neighborhood aggregator \cite{bacciu2020gentle, bronstein2021gdl}.
We refer to this form of aggregation as \emph{local}, as opposed to \emph{non-local} aggregation where \aggr runs over nodes outside of $N(u)$, \eg by means of virtual nodes \cite{sestak2024vnegnn}.
In node classification, the top-layer node embeddings are stacked into a matrix $H^L \in \bbR^{n \times d}$, which is fed to a dense layer to obtain a label distribution $p_\theta(\vY \mid \GX) = \mathrm{softmax}(H^L W)$, where $W \in \bbR^{d \times c}$ are weights.
In graph classification, they are aggregated into an overall embedding $\vh_G = \aggr_G(\{\!\{ \vh_u^L : u \in V \}\!\})$, also fed to a dense layer.

\textbf{Explanations and Faithfulness.}  There exist a number of post-hoc techniques \citep{longa2022explaining, agarwal2023evaluating, kakkad2023survey} that given a GNN $p_\theta$ can extract a \textit{local explanation} for any target decision $(\GX, \hat{y})$.  These explanations identify a subgraph $\RX$ of $\GX$ capturing those elements -- edges and/or features -- deemed relevant for said decision.
Unfortunately, explanations output by post-hoc approaches may fail to identify all and only the truly relevant elements,\footnote{The tacit assumption in the XAI and domain invariance literature is that these elements exists.} in which case we say they are not \textit{faithful} to the GNN's reasoning process \citep{longa2022explaining, agarwal2023evaluating}.  Lack of faithfulness hinders understanding, trust allocation, and debugging \citep{teso2023leveraging}.

This has prompted the development of \textbf{\textit{self-explainable GNNs}} (\SEGNNs), a class of GNNs that natively  output explanations without any post-hoc analysis \citep{kakkad2023survey, christiansen2023faithful}.
\SEGNNs comprise two modules:  the \textit{detector} $\DET$ extracts a class-discriminative subgraph $\RX$ from the input, and the \textit{classifier} $\CLF$ uses $\RX$, and only $\RX$, to infer a prediction.  In this context, $\RX$ acts as a local explanation.
Both modules are GNNs, and the detector is encouraged to output human interpretable explanations by leveraging attention \cite{miao2022interpretable, lin2020graph, serra2022learning, wu2022discovering}, high-level concepts or prototypes \citep{zhang2022protgnn, ragno2022prototype, dai2021towards, dai2022towards, magister2022encoding}, or other techniques \cite{yu2020graph, yu2022improving, miao2022interpretablerandom, giunchiglia2022towards}.

In practice, the detector outputs per-element relevance scores (usually in $[0, 1]$), which are then used to identify an explanation subgraph $\RX$ via, \eg thresholding.  Importantly, the node features $X^R \in \bbR^{|\RX| \times d}$ of $\RX$ are derived from, but \textit{not necessarily identical to}, those of $\GX$.  These remarks will become relevant when discussing strategies for improving faithfulness in \cref{sec:perfect_faith_modular}.

\textbf{Domain Invariance.}  \textbf{\textit{Domain-invariant GNNs}} (\DIGNNs) aim to generalize across related domains \cite{cai2023learning, chen2023rethinking, gui2023joint, jiang2023survey}.
The underlying assumption is that the label $y$ depends only on (hidden) \textit{domain-invariant} factors, while the input graph $\GX$ is contaminated by domain-dependent elements.  The literature suggests that, in order to generalize across domains, \DIGNNs have to be \textit{\textbf{plausible}}, that is, they should recover the (unobserved) truly invariant subgraph as well as possible \cite{jiang2023survey, schölkopf2021causal}.

Like \SEGNNs, \DIGNNs are modular, \ie they pair a detector $\DET$ for identifying the invariant subgraph $\RX$ with a classifier $\CLF$ taking $\RX$ as input.  The detector is encouraged to output highly plausible subgraphs through specialized regularization terms and training strategies \citep{chen2023rethinking, gui2023joint, chen2022learning, li2022learning}.
We will show in \cref{sec:sufficiency-invariance} that this is not enough to ensure the model as a whole is domain invariant.

\section{Are all Faithfulness Metrics the Same?}
\label{sec:faithfulness}

In the remainder, we fix a GNN $p_\theta$ and a decision $(\GX, \hat y)$ and study the faithfulness of a corresponding local explanation $\RX$.
Intuitively, an explanation is faithful insofar as it is \textit{sufficient}, \ie keeping it fixed shields the model's output from changes to its complement $\CX$, and \textit{necessary}, \ie altering it affects the model's output even if the complement $\CX$ is kept fixed \citep{watson2021local, beckers2022causal, marques2022delivering, darwiche2023complete}.
Formally, we say that an explanation $\RX$ is \textbf{\textit{strictly sufficient}} if no change to the complement $\CX$ induces any change to the model's output, and \textbf{\textit{strictly necessary}} if all changes to the explanation lead to a change.  A \textit{\textbf{strictly faithful}} explanation satisfies both conditions.

Existing metrics make these notions practical by restricting the set of perturbed graphs they consider and by defining a relaxed measure of expected change in model output \citep{sanchez2020evaluating}.
Specifically, \textit{unfaithfulness} ($\UNF$) \citep{agarwal2023evaluating} estimates sufficiency as the Kullback-Leibler divergence between the original label distribution and that obtained after zeroing-out all irrelevant features from $\GX$.
\textit{Fidelity minus} ($\FIDM$) \citep{pope2019explainability, yuan2022explainability, amara2022graphframex} instead erases all edges and features deemed irrelevant by the explanation and measures the change in likelihood of the prediction $\hat y$.\footnote{A variant based on the difference in accuracy also exists \citep{yuan2022explainability}.}
\textit{Robust fidelity minus} ($\RFIDM$) does the same but repeatedly deletes random edges from $\CX$ \citep{Zhao2022consistent, amara2023ginxeval, zheng2023robust}.
Finally, \textit{probability of sufficiency} ($\PROBSUFF$) estimates how often the model's prediction changes after multiplying the node features with the relevance scores output by the detector \citep{Juntao2022CF2}.
%
%
Metrics for necessity are specular, \ie they manipulate the input by removing \textit{relevant} elements instead, and include \textit{fidelity plus} ($\FIDP$), \textit{robust} fidelity plus ($\RFIDP$), and \textit{probability of necessity} ($\PROBNEC$).

\begin{table}[!t]
    \centering
    \caption{\textbf{\cref{def:suf-nec} recovers existing faithfulness metrics} for approriate choices of divergence $d$ and interventional distributions $\PR$ and $\PC$.}
    \label{tab:list-of-metrics}
    \small
    \scalebox{0.875}{
    \begin{tabular}{llcl}
        \toprule
        \textbf{Metric}
            & \textbf{Estimates}
            & \textbf{Divergence $d$}
            & \correction{\textbf{Distribution $q$}}{\textbf{Allowed changes}}
        \\
        \midrule
        \UNF \citep{agarwal2023evaluating}
            & \SUF
            & $\KL(p_\theta(\cdot \mid \GX), p_\theta(\cdot \mid \GX'))$
            & zero out all irrelevant features
        \\
        \FIDM \citep{pope2019explainability, yuan2022explainability, amara2022graphframex}
            &
            & $| p_\theta( \hat y \mid \GX) - p_\theta( \hat y \mid \GX') |$
            & zero out all irrelevant features, delete all irrelevant edges
        \\
        \RFIDM \citep{Zhao2022consistent, amara2023ginxeval, zheng2023robust}
            &
            & \correction{"}{$\bbE | p_\theta( \hat y \mid \GX) - p_\theta( \hat y \mid \GX') |$}
            & delete a random subset of irrelevant edges
        \\
        \PROBSUFF  \citep{Juntao2022CF2}
            &
            & $\Ind{ p_\theta( \hat{y} \mid \GX) = p_\theta( \hat{y} \mid \GX') }$
            & multiply all irrelevant elements by relevance scores
        \\
        \midrule
        \FIDP \citep{pope2019explainability, yuan2022explainability, amara2022graphframex}
            & \NEC
            & $| p_\theta( \hat y \mid \GX) - p_\theta( \hat y \mid \GX') |$
            & zero out all relevant features, delete all relevant edges
        \\
        \RFIDP \citep{Zhao2022consistent, amara2023ginxeval, zheng2023robust}
            &
            & \correction{"}{$\bbE | p_\theta( \hat y \mid \GX) - p_\theta( \hat y \mid \GX') |$}
            & delete a random subset of relevant edges
        \\
        \PROBNEC  \citep{Juntao2022CF2}
            &
            & $\Ind{ p_\theta( \hat{y} \mid \GX) \ne p_\theta( \hat{y} \mid \GX') }$
            & multiply all relevant elements by relevance scores
        \\
        \bottomrule
    \end{tabular}
    }
\end{table}

While existing metrics differ in many details, they nicely fit the same common format:

\begin{defn}
    \label{def:suf-nec}
    Let $d$ be a divergence, $\PR$ a distribution over supergraphs of $\RX$, and $\PC$ a distribution over supergraphs of $\CX$.
    Also, let $\Delta_d(\GX, \GX') = d( p_\theta( \cdot \mid \GX) \ \| \ p_\theta( \cdot \mid \GX' ) )$
    measure the impact of replacing $\GX$ with $\GX'$ on the label distribution.\footnote{We drop the dependency on $\hat y$ for brevity.}
    Then, the \textbf{\textit{degree of sufficiency}} and \textbf{\textit{degree of necessity}} of an explanation $\RX$ for a decision $(\GX, \hat y)$ are:
    \[
        \SUF_{d, \PR}(\RX) =
            \mathbb{E}_{\GX' \sim \PR} [
                \Delta_d(\GX, \GX')
            ],
        \quad
        \NEC_{d, \PC}(\RX) =
            \mathbb{E}_{\GX' \sim \PC} [
                \Delta_d(\GX, \GX')
            ]
        \label{eq:suf-nec}
    \]
\end{defn}

The expectations in \cref{eq:suf-nec} are potentially unbounded, but can be \textit{normalized} to $[0, 1]$, the higher the better, via a non-linear transformation, \ie taking $\exp( - \SUF_{d,\PR}(\RX))$ and $1 - \exp(-\NEC_{d,\PC}(\RX))$.
\cref{tab:list-of-metrics} shows that \cref{def:suf-nec} recovers all existing metrics for appropriate choices of divergence $d$ and distributions $\PR$ and $\PC$.
Alternative choices immediately yield additional metrics.

While \textit{both sufficiency and necessity matter, there exists a natural tension between them}.  For instance, explanations covering a larger portion of $\GX$ likely attain higher sufficiency but lower necessity.
This motivates us to define the \textbf{\textit{degree of faithfulness}} $\FAITH_{d,\PR,\PC}(\RX)$ as the harmonic mean of normalized sufficiency and necessity, which is biased towards the lower of the two.
We henceforth suppress the dependency on $d$, $\PR$ and $\PC$ whenever it is clear from context.

\subsection{Faithfulness metrics are not interchangable}
\label{sec:metrics-are-not-interchangeable}

A first key observation is that, despite falling into a common template, existing metrics are \textit{not interchangeable}, in the sense that explanations that are highly faithful according to one metric can be arbitrarily unfaithful for the others.
We formalize this in the following proposition:

\begin{restatable}{proposition}{propmetricsnotequiv}
    \label{prop:metrics-are-not-equivalent}
    \textit{(Informal.)} Let $(\PR, \PC)$ be a pair of distributions as per \cref{def:suf-nec}. Then, depending on $p_\theta$ and $\GX$, it is possible to find $(\PR', \PC')$ such that $|\SUF_{d,\PR}(\RX) - \SUF_{d,\PR'}(\RX)|$ and $|\NEC_{d,\PC}(\RX) - \NEC_{d,\PC'}(\RX)|$ are as large as the natural range of $d$.
\end{restatable}

All proofs (as well as relevant discussion) can be found in \cref{sec:proofs}.
In essence, \cref{prop:metrics-are-not-equivalent} means that faithfulness results cannot be properly interpreted unless the parameters $d$, $\PR$ and $\PC$ are known.
In practice, this entails that explanation producers -- such as banks and algorithm designers -- responsible for certifying the faithfulness of explanations cannot withhold the parameters used in the computation, lest end-users blindly trust explanations that are not sufficiently faithful for their downstream applications.

\subsection{Not all faithfulness estimators are equally good}
\label{sec:estimators_not_equally_good}

Given this large variety of metrics, it is natural to wonder whether they are all equally good.
As we will show, it turns out this is not the case because -- surprisingly -- commonly used necessity metrics are \textit{insensitive} to the number of irrelevant edges in the explanation.

\textbf{\FIDP is systematically invariant to truly irrelevant edges.}  In the following, we assume $\RX$ contains $m$ \textit{truly relevant edges}, that is, such that if any of them is removed from $G$ yield a new graph $G'$ such that $\Delta(G, G') \ne 0$.
We begin by showing that metrics that estimate necessity using a single modified graph $\GX'$ obtained by deleting the whole explanation $\RX$,\footnote{In node classification, the node whose prediction is being explained is not removed even if it belongs to $R$.} such as \FIDP, suffer from this issue.
To see this, note that in an $L$-layer GNN with local \aggr, messages from nodes outside of $N_L(u)$ cannot influence the prediction of node $u$.
Now, consider a 1-layer GNN $p_\theta$ for node classification and an input line graph $u_1 \leftarrow u_2 \leftarrow \ldots \leftarrow u_n$.
Since there is only one layer, only the messages from $N_1(u_1) = \{ u_1, u_2 \}$ can contribute to the distribution of $Y_1$.
However, \FIDP associates the same value to both $\RX = N_2(u_1)$ and to $\RX' = \GX$, which contains arbitrarily many irrelevant nodes.
In fact, for $\RX$ it deletes $u_2$, while for $\RX'$ it deletes $u_2, \ldots, u_n$:  in both cases, the model predicts $Y_1$ using $\vx_1$ only, meaning that \FIDP will attain the same value.
Note that, when the edge relevance scores are binary, \PROBNEC behaves exactly like \FIDP, in which case it also suffers from this issue.

\textbf{\RFIDP is systematically invariant to truly irrelevant edges.}  So does \RFIDP.  To see this, consider a 1-layer GNN for node prediction, and let $\GX$ be an $n$-node star graph with $u_1$ as its center, and assume that only the edges between $u_1$, $u_2$ and $u_3$ are truly relevant to $p_\theta(Y_1 \mid \GX)$.  Now, take two explanations: $R = \{u_1, u_2, u_3\}$ covers all and only the truly relevant edges, and $R' = G$.  Like above, it turns out that $\RFIDP(R) = \RFIDP(R')$.  In fact, \RFIDP estimates necessity by removing edges in an IID fashion, which means that the probability that it deletes the edges between $u_1$, $u_2$, and $u_3$ from $R$ is the same regardless of how many other edges appear in the explanation.
\correction{We formalize this intuition in the following result:}{We provide an illustrative figure of the previous two examples in \cref{app:example_illustration} and formalize the intuition above in the following result:}

\begin{restatable}{proposition}{proprfidexpval}
    \label{prop:rfid_expval}
    Fix a divergence $d$ and a threshold $\epsilon > 0$.  Let $R$ contain $r$ truly relevant edges.  Then, $\RFIDP(R)$ does not depend on $\norm{R} - r$.
\end{restatable}

The question is then how to instantiate \NEC such that it \textit{does} depend on the number of irrelevant elements.
We next show that this can be achieved by letting \PC be the \textit{uniform distribution} over all subgraphs of $\RX$.
Since marginalizing over all subgraphs is intractable, we also show that simply averaging over a smaller subset of graphs obtained by erasing $\budget$ edges from the explanation itself also avoids the issue.
We summarize our findings as follows:


\begin{restatable}{proposition}{propnecexpval}
    \label{prop:nec_expval}
    Fix any divergence $d$ and a constant budget $\budget \ge 1$.
    Let $\calS_R^\budget$ be the set of subgraphs of $G$ obtained by deleting $\budget$ edges from $R$ while keeping $C$ fixed, and $\calA_R^\budget = \{ G' \in \calS_R^\budget : \Delta(G, G') \ge \epsilon \}$ those subgraphs that lead to a large enough change in $\Delta$.  Also, let $\calS_R = \calS_R^1 \cup \ldots \cup \calS_R^m$ and similarly for $\calA_R$.
    Given an explanation $R$ containing $r$ truly relevant edges, $\NEC(R)$ computed using a uniform $\PC$ over $\calS_R$ depends on the number of irrelevant edges $\norm{R} - r$.
    The same holds if $\NEC(R)$ is computed using a uniform $\PC$ over $\calS_R^\budget$.
\end{restatable}

This result highlights that, by selecting an appropriate interventional distribution $\PC$, it is possible to design a necessity metric that appropriately accounts for the number of irrelevant edges.

\textbf{The choice of budget $\budget$ can bias necessity.}  In fact, \cref{prop:nec_expval} may fail if \budget depends on the size of the input graph or the size of the explanation.  This occurs because the numerator and denominator of $|\calA_R^\budget|/|\calS_R^\budget|$ both depend on \budget, and thus can act as a confounding variable.
Specifically, if \budget depends on $\norm{R}$, the metric -- once again -- does not properly account for irrelevant items in the explanation, as the number of deletions increases according to the size of the explanation. 
Indeed, a numerical simulation shows that the probability of deleting at least one truly relevant edge does not depend on the number of irrelevant ones in the explanation, if \budget is proportional to the explanation size 
(see~\cref{fig:budget-and-probability} in the Appendix).
%
On the other hand, if \budget is proportional to the size of the input graph $G$, the number of deletions depends also on the size of the complement, resulting in a metric confounded by the complement of the explanation\footnote{\correction{From a causal perspective, letting \budget depend on the size of $G$ opens a backdoor path from $G$ to \NEC, confounding \NEC .}{From a causal perspective, letting \budget depend on the size of $G$ entails the causal graph $\NEC \leftarrow \RX \leftarrow G \rightarrow \budget \rightarrow \NEC$ which contains the backdoor path $\RX \leftarrow G \rightarrow \budget \rightarrow \NEC$, confounding the estimation of \NEC.}\cite{pearl2009causality}}.
\correction{}{\ADVROB suffers from this issue \cite{fang2023evaluating}.}
This is best seen in the following example:

\begin{example}
    Fix an explanation $R$ composed of two edges, only one of which is truly relevant.
    If $G$ has $10$ edges, fixing a budget of deletion as $10\%$ of the graph size yields a modified graph $G'$ such that $P(G' \in \calA^1_R) = \frac{1}{2}$.
    Instead, if $G$ has $20$ edges, then $P(G' \in \calA^2_R) = 1$.
    Hence, despite $R$ being fixed, the probability of sampling truly relevant edges drastically changes.
\end{example}

We propose to avoid these issues by fixing a budget that is proportional to a data set-wide statistic, such as average graph size $\bar{m} = 1/|\calD| \sum_{G \in \calD} \norm{G}$, where $\calD$ is the set of graphs.  This way, it no longer directly depends on the size of any specific input graph or explanation, while still being adaptive to the target task.

\begin{figure}[!t]  
    \floatbox[{\capbeside\thisfloatsetup{capbesideposition={left,top},capbesidewidth=.49\textwidth}}]{figure}[\FBwidth]
    {\caption{
    \textbf{Dependency of the necessity metrics on the size of the explanation.}
    \RFIDP and \NEC of explanations output by \LECI on \MotifTwoBasis (averaged over 5 seeds) for different explanation sizes (x-axis) and metric hyper-parameter $\kappa, \budget \in \{ 3\%, 5\% \}$. \RFIDP assigns similar or even higher scores to larger explanations, while \NEC (with a budget $\budget$ proportional to the average graph size $\bar{m}$) decreases for increasing explanation size, as expected.}
    \label{fig:budget-and-insensitivity}}
{\includegraphics[width=0.45\textwidth]{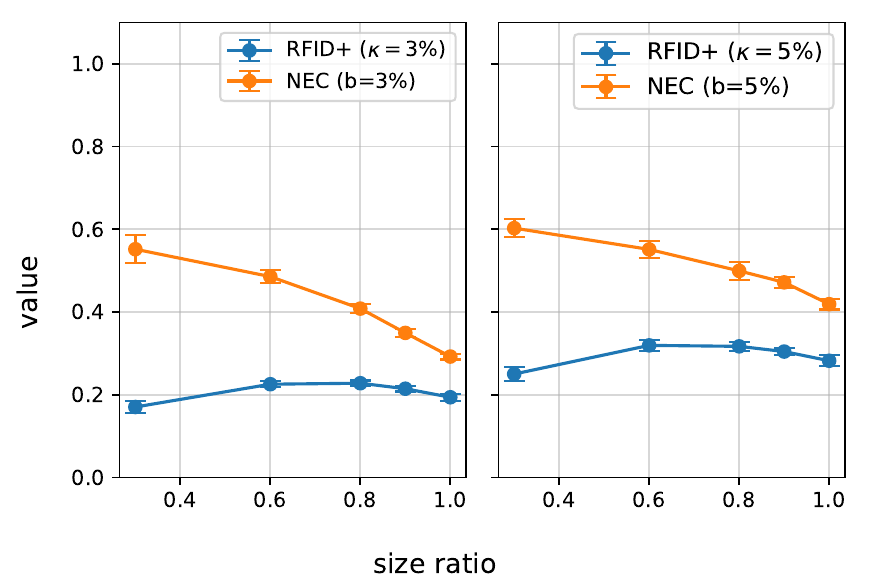}}
\end{figure}

Based on these observations, in our experiments we adopt a uniform $\PC(G)$ over subgraphs of size $\norm{G} - \budget$, where \budget is a hyper-parameter, which will be denoted as $\PC^\budget(G)$.
The practical difference between \RFIDP and \NEC with $\PC^\budget$ is exemplified in the following analysis, where we evaluate the necessity of explanations of the modular architecture \GSAT \cite{miao2022interpretable} over the benchmark \MotifTwoBasis \cite{gui2022good}, and for different \RFIDP edge-deletion probability $\kappa$ and budget \budget. Since \GSAT outputs explanation scores in the $[0,1]$ range, we obtain a relevant input subgraph by discretizing the explanation mask at different size ratios, as common in XAI evaluation \cite{longa2022explaining, amara2022graphframex}. The results provided in \cref{fig:budget-and-insensitivity} show that \RFIDP assigns close-to-constants scores to explanations at different size ratios, while our previously discussed \NEC tends to penalize larger explanations. We provide a further comparison of the two metrics in \cref{app:further_experiments_nec}.

\section{Is Faithfulness Worth Optimizing For?}
\label{sec:perfect-faith}

We show that for regular injective GNNs, strict faithful explanations are trivial, highlighting a trade-off between explainability and expressivity. 
For modular GNNs, instead, non-trivial strictly faithful explanations are \correction{possible}{theoretically attainable} provided certain conditions are met \correction{}{(see \cref{app:strict_faith_modular})}. We then empirically verify that some of these conditions are hard to achieve for practical models, suggesting a trade-off between strict faithfulness and learnability.

\subsection{The Case of Regular GNNs}

Recall that an explanation is strictly faithful (cf. \cref{sec:faithfulness}) when no change to the complement $\CX$ of $\RX$ affects the model’s output, and when all elements in $\RX$ contribute to the final prediction.
In the following, we show that for regular GNNs strictly faithful explanations are \textit{completely uninformative}.
To build intuition, notice that strictly sufficient explanations must subsume the computational graph of the prediction, that is, the subgraph of $\GX$ induced by all nodes whose messages influence the label distribution.  In node classification, this is the $L$-hop neighborhood of the target node, and in graph classification the whole input graph.
More formally:

\begin{restatable}{proposition}{propssearetrivial}
    \label{prop:sse-are-trivial}
    Consider a binary classification task and an $L$-layer injective GNN:
    \textit{i}) for \textbf{node classification} that only uses local aggregators.  Then an explanation $\RX$ for a decision $(\GX, \hat y_u)$ is strictly faithful wrt \PR iff it subsumes $N_L(u)$.
    \textit{ii}) For \textbf{graph classification}.  Then an explanation $\RX$ for a decision $(\GX, \hat y)$ is strictly faithful wrt \PR iff it subsumes all of $\GX$.
\end{restatable}

Note that in both cases the explanation does not depend on the learned weights at all, hence strictly faithful explanations are \textit{trivial}. This is a direct consequence of injectivity, a prerequisite for GNNs to implement the Weisfeiler-Lehman test \citep{xu2018powerful, bianchi2024expressive}, and highlights a trade-off between expressivity and explainability.\footnote{Of course, one can obtain more informative explanations by giving up on strict faithfulness and instead aiming at producing explanations that are both faithful enough and smaller than the entire computational graph.} For regular GNNs, even forgoing injectivity does not give an easy way to obtain non-trivial strictly faithful explanations. In the next section, we show how this is instead attainable with modular GNN architectures.

\subsection{The Case of Modular GNNs}
\label{sec:perfect_faith_modular}

Rather counter-intuitively, despite modular GNNs being designed to ensure predictions to adhere the underlying explanation, it has been shown that popular \SEGNNs suffer from faithfulness issues \cite{christiansen2023faithful}.
To better understand this phenomenon, we highlight four common architectural design choices whose absence easily allows information leakage from the complement to the explanation, showing how popular \SEGNNs and \DIGNNs systematically fail in fulfilling all of our desiderata.  Specifically:
\begin{itemize}[leftmargin=1.25em]

    \item \textbf{Hard Scores} (\textbf{HS}): \DET should put exact zero importance to information in the complement, otherwise \upd and \aggr will inevitably mix information of \RX and \CX;  

    \item  \textbf{Content Features} (\textbf{CF}): since the message passing of \DET is unconditional on its predicted edge relevance score, the classifier \CLF should have access to raw input (or content) features, contrarily to using the detector's latent representation which is heavily leaked;  

    \item \textbf{Explanation Readout} (\textbf{ER}): (for graph classification only) the final graph global readout should aggregate only over \RX. Since explanations are oftentimes soft edge masks over the entire graph, scaling the node embedding according to their average importance scores can still reduce the impact of irrelevant nodes; 

\end{itemize}

The previous architectural desiderata are all tackling the issue of the classifier not correctly using \RX for making predictions. 
\textit{\textbf{This is however not sufficient for a perfectly faithful model}}, and a rather complementary direction of investigation is that of making the detector \DET more stable to irrelevant modifications. 
In fact, assuming a classifier perfectly implementing the desiderata above and an explanation \RX, it is easy to see that if the detector changes the predicted explanation for a modified graph $\GX' \supseteq \RX$, then the final prediction can arbitrarily change, thus also \SUF does.
%
Following on this, we can outline another widespread architectural design choice playing a direct role in how the detector produces explanations:
\begin{itemize}[leftmargin=1.25em]

    \item  \textbf{Local Aggregations} (\textbf{LA}): non-local aggregations (cf. \cref{sec:background}) can easily mix the information of the explanation with that of its complement, and can create unwanted dependencies between any pair of nodes in the graph. \SA{LA, in principle, can only help to improve a bit the stability, as it permits the model to be insensible to far away perturbations. However, the model can still be unstable inside its computational graph. Integrate when we have more results on stability with LA}
    
\end{itemize}

In the remainder of this section, we provide an analysis on the stability of popular modular GNNs and empirically investigate the impact of integrating them with the strategies they lack, measuring their effect on both faithfulness and final accuracy \SA{and stability?}.

\begin{wrapfigure}{r}{0.4\textwidth}
    \begin{minipage}{\linewidth}
     \captionof{table}{\textbf{Popular modular GNNs fail to fully implement faithfulness-enforcing strategies}. \SEGNNs at the top and \DIGNNs on the bottom. \xmark/\cmark means that both variants exist and the choice is made via cross-validation.} 
        \small
        \begin{tabular}{lccccc}
        \toprule
        \textbf{Model} & \textbf{HS} & \textbf{ER} & \textbf{CF} & \textbf{LA}\\
        \midrule
         \GISST \cite{lin2020graph}   & \xmark  & \xmark & \cmark & \cmark\\
         \GSAT \citep{miao2022interpretable} & \xmark & \xmark & \cmark & \xmark/\cmark\\
         \RAGE \citep{kosan2023robust}   & \xmark  & \xmark & \cmark & \cmark\\    
        \midrule
         \CIGA \citep{chen2022learning} & TopK & \cmark & \xmark/\cmark & \xmark/\cmark\\
         \GSAT \citep{miao2022interpretable} & \xmark & \xmark & \cmark & \xmark/\cmark\\
         \LECI \citep{gui2023joint} & \xmark & \xmark & \cmark & \xmark/\cmark\\
        \bottomrule
        \end{tabular}
        \label{tab:taxonomy_tmp}
    \end{minipage}
\end{wrapfigure}

\textbf{Experimental setup}:
We benchmarked six representative modular architectures, listed in \cref{tab:taxonomy_tmp}, following their respective evaluation testbed for graph classification.
The datasets chosen for evaluation are picked from the usual evaluation routines for respectively \SEGNNs and \DIGNNs. For \DIGNNs, the chosen benchmarks can be divided into datasets with known domain-invariant input motifs (\MotifTwoBasis, \CMNIST), datasets with presumed domain-invariant input motifs (\LBAPS), and datasets where domain-invariant input motif are not expected (\SSTTwoS). This last case is especially problematic, as no clear advantage is expected in focusing on a subset of the input graph. 
The full experimental setup is detailed in \cref{app:experimental_details}.

\textbf{Results and discussion}:
\cref{tab:mitigations_SE} and \cref{tab:mitigations_DI} show the results for the base \SEGNNs and \DIGNNs architectures and their augmented variants.
%
The results suggest that it is possible to improve faithfulness by applying the missing strategies above, yet not consistently across strategies, models, and datasets. Specifically, the strategy with the lightest impact on model architecture, namely \textbf{ER}, is the one achieving the best results, improving faithfulness over the base model in $12$ out of $21$ cases and being on par in 4. Conversely, more impactful strategies like \textbf{HS} and \textbf{LA} yield less consistent results. For example, \textbf{HS} severely compromises the training of \GSAT in \MotifTwoBasis and \MotifSize, resulting in a train and test accuracy around $50\%$. 
\correction{}{We provide a more detailed discussion of this issue in \cref{app:impl_details}.}
When combining strategies, faithfulness generally improves when the individual strategies yield positive results.
Among all datasets, \SSTTwoS is confirmed to be the most difficult to improve on. Indeed, models tend to output explanations that cover the entire input graph for this dataset (see 
\cref{app:failure_cases} for the details). 

\correction{These results suggest that, while potentially useful, strict faithfulness is a hard goal to achieve and not all strategies encouraging it are viable in practice, \textbf{questioning the current design of modular GNNs}.}{These results show that, despite the enforcement of simple architectural desiderata targeted to reduce the information leakage between complement and explanation, faithfulness is not consistently improving and accuracy sometimes drops.
\textbf{This questions the current design of modular GNNs}, suggesting that current popular models might be rooted in architectural components implicitly favoring leakage, and that the factors contributing to faithfulness may be more nuanced than previously presumed.
In addition, we acknowledge that the desiderata outlined above may be insufficient in addressing the full scope of the problem, highlighting that further research is needed on more advanced faithfulness-enforcing strategies.}

\correction{}{As additional evidence, we show in \cref{tab:stability} that training regimes of most GNNs indeed lead to sensibly unstable models, further hindering faithfulness and suggesting that training-time mitigations are likely to be key.}
\correction{Increasing the degree of faithfulness is a convenient strategy when strict faithfulness is out of reach.}{}
\correction{}{In fact,} necessity can intuitively be encouraged via sparsification techniques \cite{lin2020graph, kosan2023robust}, whereas encouraging stability of the detector $\DET$ (using approaches like contrastive learning \cite{chen2022learning, chen2023rethinking}, adversarial training \cite{gui2023joint}, or data augmentation \cite{wu2022discovering}) is a natural option for encouraging sufficiency, {\em regardless} of the properties of the classifier \CLF built on top of it. 
\correction{}{We plan to investigate how those insights translate into a new faithful-by-design modular architecture in future work.}

\begin{wrapfigure}{r}{0.4\textwidth}
    \begin{minipage}{\linewidth}
     \captionof{table}{\textbf{Popular modular architectures are not stable}. Matthews Correlation Coefficient (MCC) between the original binarized explanation and the binarized explanation after perturbing the topology of the original complement. A value of $1$ means perfect correlation, thus perfect stability. Overall, all models are significantly below $1$. \GSAT appears twice because of different hyper-parameters.} 
        \small
        \begin{tabular}{lcc}
        \toprule
        \textbf{Model} & \textbf{\MotifTwoBasis} & \textbf{\MotifSize} \\
        \midrule
         \GSAT   & \nentry{04}{01} & \nentry{22}{02} \\
         \GISST  &  &  \\
         \RAGE   &  &  \\
        \midrule
         \LECI   & \nentry{44}{07} & \nentry{62}{08} \\
         \CIGA   & \nentry{53}{04} & \nentry{59}{01} \\
         \GSAT   & \nentry{08}{01} & \nentry{21}{01} \\
        \bottomrule
        \end{tabular}
        \label{tab:stability}
    \end{minipage}
\end{wrapfigure}

\begin{table}
    \centering
    \caption{Test set accuracy and faithfulness of \SEGNNs augmented with the strategies delineated in \cref{sec:perfect_faith_modular}, averaged over 5 seeds. Improvements over the base model are in \textbf{\textcolor{ForestGreen}{green}}, worsenings in \textbf{\textcolor{BrickRed}{red}}.  We do not augment \GISST on \BAMULTISHAPES, \MotifTwoBasis, and \MotifSize since these data sets have constant input features for all nodes, meaning that \correction{these strategies}{the modifications} have no effect.}
    \label{tab:mitigations_SE}
    \footnotesize
    \scalebox{.9}{
    \begin{tabular}{lcccccccc}
        \toprule
         Dataset & \multicolumn{2}{c}{\BAMULTISHAPES}  & \multicolumn{2}{c}{\MotifTwoBasisS} & \multicolumn{2}{c}{\MotifSize} & \multicolumn{2}{c}{\BBBP}\\
         \cmidrule(lr){2-3} \cmidrule(lr){4-5} \cmidrule(lr){6-7} \cmidrule(lr){8-9}
         & \ACC & \FAITH & \ACC & \FAITH & \ACC & \FAITH & \ACC & \FAITH \\
         \midrule
         \GSAT                  & \nentry{100}{00}  & \nentry{35}{03} & \nentry{92}{01} & \nentry{61}{01} 
                                & \nentry{90}{01}   & \nentry{60}{02} & \nentry{79}{04} & \nentry{27}{08}\\
         \GSAT + \textbf{ER}    & \nentry{100}{00}  & \nentry{35}{03} & \nentry{92}{01} & \gentry{63}{01} 
                                & \nentry{90}{01}   & \gentry{65}{01} & \gentry{80}{02} & \gentry{33}{04}\\
         \GSAT + \textbf{HS}    & \rentry{98}{01}   & \rentry{21}{06} & \rentry{53}{02} & \rentry{24}{05} 
                                & \rentry{54}{03}   & \rentry{22}{05} & \rentry{71}{01} & \gentry{31}{09}\\
        \GSAT + \textbf{ER} + \textbf{HS} & \rentry{99}{01} & \rentry{24}{04} & \rentry{57}{04} & \rentry{37}{03} & \rentry{56}{07} & \rentry{29}{09} & \rentry{73}{02} & \gentry{33}{02}\\
        
        \midrule
         \GISST                 & \nentry{100}{00} & \nentry{25}{03} & \nentry{92}{01} & \nentry{53}{02} 
                                & \nentry{92}{00}  & \nentry{50}{02} & \nentry{84}{03} & \nentry{23}{11} \\
                                
         \GISST + \textbf{ER}   & - & - & - & - 
                                & - & - & \gentry{85}{06} & \gentry{27}{06} \\
         \GISST + \textbf{HS}   & - & - & - & - 
                                & - & - & \rentry{83}{05} & \rentry{19}{07}  \\
         \GISST + \textbf{ER} + \textbf{HS}   & - & - & - & - 
                                & - & - & \rentry{81}{07} & 
                                \rentry{15}{09} \\
        
        \midrule
         \RAGE               &  \nentry{96}{01} & \nentry{33}{05} & \nentry{83}{02} & \nentry{64}{04} & 
                                \nentry{74}{09} & \nentry{63}{07} & \nentry{82}{01} & \nentry{33}{04}\\
         \RAGE + \textbf{ER} &  \nentry{96}{02} & \nentry{33}{02} & \gentry{85}{06} & \gentry{66}{03} &
                                \rentry{71}{09} & \rentry{55}{07} & \gentry{84}{01} & \nentry{33}{05} \\
         \RAGE + \textbf{HS} &  \gentry{97}{01} & \gentry{46}{03} & \gentry{85}{01} & \gentry{65}{02} & 
                                \gentry{78}{07} & \gentry{65}{09} & \gentry{84}{02} & \gentry{46}{02} \\
        \RAGE + \textbf{ER} + 
                \textbf{HS}  &  \nentry{96}{01} & \gentry{46}{04} & \nentry{83}{04} & \nentry{64}{04}
                             &  \gentry{75}{08} & \rentry{62}{12} & \nentry{82}{01} & \gentry{43}{03}\\ 
        
        \bottomrule
    \end{tabular}
    }
\end{table}

\begin{table}
    \centering
    \footnotesize
    \caption{OOD test set accuracy and faithfulness of \DIGNNs augmented with the strategies delineated in \cref{sec:perfect_faith_modular}, averaged over 5 seeds.}
    \label{tab:mitigations_DI}
    \scalebox{.9}{
    \begin{tabular}{lcccccccc}
        \toprule
         Dataset & \multicolumn{2}{c}{\MotifTwoBasisS}  & \multicolumn{2}{c}{\CMNISTS} & \multicolumn{2}{c}{\LBAPS} & \multicolumn{2}{c}{\SSTTwoS}\\
         
         \cmidrule(lr){2-3} \cmidrule(lr){4-5} \cmidrule(lr){6-7} \cmidrule(lr){8-9}
         & \ACC & \FAITH & \ACC & \FAITH & \ACC & \FAITH & \ACC & \FAITH \\
        \midrule
         \LECI                & \nentry{85}{07} & \nentry{58}{02} & \nentry{26}{10} & \nentry{48}{11} 
                              & \nentry{71}{01} & \nentry{43}{05} & \nentry{83}{01} & \nentry{26}{02}\\
         \LECI + \textbf{ER}  & \gentry{86}{03} & \gentry{59}{02} & \gentry{58}{12} & \gentry{58}{03} 
                              & \nentry{71}{01} & \gentry{46}{02} & \rentry{82}{01} & \rentry{13}{02}\\
         \LECI + \textbf{HS}  & \gentry{86}{04} & \rentry{57}{02} & \gentry{34}{10} & \gentry{57}{01} 
                              & \gentry{72}{01} & \rentry{24}{01} & \nentry{83}{01} & \rentry{18}{03}\\
         \LECI + \textbf{LA}  & - & -                             & \gentry{46}{11} & \gentry{57}{03} & \rentry{69}{01} 
                              & \rentry{31}{03} & \rentry{81}{05} & \rentry{20}{04}\\
        \LECI + \textbf{ER} + \textbf{HS} + \textbf{LA} & \rentry{79}{11} & \rentry{55}{02} & \gentry{75}{06} & \gentry{61}{01} & \rentry{59}{02} & \rentry{21}{01} & \rentry{81}{02} & \rentry{16}{01}\\
        \midrule
        
         \CIGA               & \nentry{46}{10} & \nentry{38}{08} & \nentry{23}{03} & \nentry{36}{03} 
                             & \nentry{69}{01} & \nentry{33}{02} & \nentry{76}{06} & \nentry{18}{01}\\
         \CIGA + \textbf{ER} & \rentry{45}{09} & \gentry{54}{04} & \nentry{23}{02} & \gentry{43}{07}
                             & \rentry{59}{07} & \rentry{09}{06} & \rentry{74}{03} & \rentry{16}{01}\\
         \CIGA + \textbf{CF} & \gentry{53}{07} & \gentry{49}{02} & \rentry{13}{01} & \gentry{49}{09} 
                             & \rentry{49}{12} & \rentry{03}{01} & \rentry{55}{07} & \rentry{09}{12}\\
         \CIGA + \textbf{LA} & - & - & \gentry{30}{09} & \gentry{41}{03} 
                             & \rentry{68}{01} & \gentry{34}{08} & \gentry{79}{03} & \rentry{16}{01}\\ 
        \CIGA + \textbf{ER} + \textbf{CF} + \textbf{LA} & \gentry{47}{08} & \gentry{39}{07} & \nentry{23}{03} & \gentry{50}{02} & \rentry{66}{01} &         
                             \rentry{23}{10} & \nentry{76}{08} & \rentry{15}{02}\\
        \midrule
        
         \GSAT               & \nentry{75}{06} & \nentry{58}{01} & \nentry{25}{04} & \nentry{48}{03}
                             & \nentry{70}{03} & \nentry{40}{06} & \nentry{79}{04} & \nentry{22}{04}\\
         \GSAT + \textbf{ER} & \rentry{59}{06} & \gentry{60}{06} & \gentry{30}{06} & \nentry{48}{01} 
                             & \rentry{67}{01} & \rentry{32}{03} & \gentry{81}{01} & \gentry{23}{04} \\
         \GSAT + \textbf{HS} & \gentry{86}{03} & \rentry{42}{07} & \rentry{14}{03} & \rentry{44}{04} 
                             & \gentry{71}{01} & \rentry{28}{01} & \gentry{80}{02} & \rentry{17}{02}\\
         \GSAT + \textbf{LA} & - & - & \gentry{27}{03} & \rentry{46}{04} 
                             & \rentry{69}{03} & \gentry{44}{02} & \gentry{81}{01} & \rentry{21}{04}\\ 
         \GSAT + \textbf{ER} + \textbf{HS} + \textbf{LA} & \rentry{64}{08} & \rentry{47}{03} & \rentry{17}{02} & \gentry{51}{05} & \nentry{70}{01} & \rentry{38}{03} & \gentry{80}{01} & \rentry{17}{01}\\
         \bottomrule
    \end{tabular}
    }
\end{table}

Finally, results in \cref{tab:mitigations_DI} show instances of \DIGNNs where faithfulness-enforcing strategies also 
improved OOD generalization abilities. 
In the next section we formally investigate the relationship between faithfulness and OOD generalization.

\section{Faithfulness is Key for Domain Invariance}
\label{sec:sufficiency-invariance}

Finally, we study the impact of faithfulness -- and lack thereof -- on domain invariance.
In the domain invariance literature \citep{gui2023joint, chen2022learning}, one distinguishes between in-distribution (ID) and out-of-distribution (OOD) data, sampled from $p^{id}(\GX, Y)$ and $p^{ood}(\GX, Y)$, respectively, and the goal is to learn a \DIGNN $p_\theta$ that generalizes from ID to OOD data.
\correction{}{Input graphs are assumed to be comprised of an invariant $R_A^* \subseteq G_A$ and a spurious subgraph, and that the ground-truth label only depends on the invariant subgraph in a domain-invariant manner, that is, $p^{id}(Y \mid R_A^* ) = p^{ood}(Y \mid R_A^* )$ \cite{cai2023learning, chen2023rethinking, gui2023joint, jiang2023survey}. 
The degree of invariance (variance) of the predicted subgraph $R_A$ is then proportional to its \textbf{plausibility} (un-plausibility), i.e., how closely it matches $R_A^*$.}
We begin by showing that, even if the graphs $\RX$ extracted by a \DIGNN (cf. \cref{sec:background}) are \correction{domain invariant}{maximally plausible}, unless they are also strictly sufficient, the model's prediction is \textit{not} domain invariant.
\correction{}{To build intuition, take a DI-GNN that, for some input, outputs a perfectly invariant explanation $R_A$. Now, if the explanation is not strictly sufficient, by definition there exists a modification to the complement of $R_A$ (which is domain-dependent) that alters the predicted class distribution. Thus the model as a whole cannot be domain-invariant.}

\begin{restatable}{proposition}{suffdignns}
    \label{prop:suff_dignns}
    Let $\MONO $ be a modular \DIGNN such that the detector $\DET$ outputs graphs $\RX$ that are maximally \correction{domain-invariant}{plausible}, \ie that comprise all and only those elements that are constant across the target domains.
    If $\RX$ is not strictly sufficient, then the prediction is not domain invariant.
\end{restatable}

This result is significant because \DIGNNs typically optimize only for \correction{domain}{the} invariance of the extracted subgraph $\RX$, neglecting how the classifier on top then uses this.
The next result highlights that the degree of faithfulness of $\RX$ -- and specifically, that of sufficiency -- plays a direct role in ensuring the model predictions are in fact domain invariant.
\correction{}{In the following Theorem we are bounding the difference in likelihood between ID and OOD as a proxy to measure the change in the fit of the data, so that a low difference corresponds to a consistent model behavior across domains. Despite many generalisation bounds available in literature \cite{redko2020survey}, ours pertains specifically to explanation quality, both in terms of degree of invariance and degree of sufficiency.}

\begin{restatable}{theorem}{faithoodupperbound}
\label{thm:faith_ood_upper_bound}
(Informal) Let $\MONO$ be a deterministic \DIGNN with detector \DET and classifier \CLF, and $p^{id}(\GX, Y)$ and $p^{ood}(\GX, Y)$ be the ID and OOD empirical distributions, respectively.  Then:
\begin{align}
    \label{eq:faith_ood_upper_bound}
        \Big | \expectedGy{p^{id}} [\MONO(y \mid \GX)] \ - \ & \expectedGy{p^{ood}} [\MONO(y \mid \GX)] \Big | 
    \\
    & \le 
        \bbE_{\GXInv} \left[
            k_1 (\LAMBDATOPO^{id}+\LAMBDATOPO^{ood}) + k_2 (\LAMBDAFEAT^{id}+\LAMBDAFEAT^{ood}) + (\LAMBDASUFF^{id} + \LAMBDASUFF^{ood}) \right]. \nonumber
\end{align}
Here, $\bbE_{\GXInv}$ runs over all possible maximally invariant subgraphs, $k_1, k_2 > 0$ are Lipschitz constants of \MONO, $\LAMBDATOPO^{id}$, $\LAMBDATOPO^{id}$ $\LAMBDAFEAT^{ood}$, and $\LAMBDAFEAT^{ood}$ measure the \correction{(negated) degree of domain invariance}{un-plausibility} of the detector's output (with respect to topology and features, respectively), and $\LAMBDASUFF^{id}, \LAMBDASUFF^{ood}$ are the degree of sufficiency for ID and OOD data.
\end{restatable}

\correction{}{In essence, \cref{thm:faith_ood_upper_bound} shows that a DI-GNN that fits well the ID data \correction{can only}{will} fit well the OOD data if i) $R_A$ is plausible (low $\lambda_{topo}$ and $\lambda_{feat}$, thus invariant across domains), and ii) $R_A$ is highly sufficient (low $\lambda_{suff}$). The practical relevance of this result is twofold: It suggests that researchers should also consider sufficiency when evaluating DI-GNNs, as well as focus on invariance of both topology and feature, rather than topology only as often done \cite{gui2023joint}.}

\begin{wrapfigure}{r}{0.40\textwidth}
    \vspace{-1em}
    \includegraphics[width=0.85\linewidth]{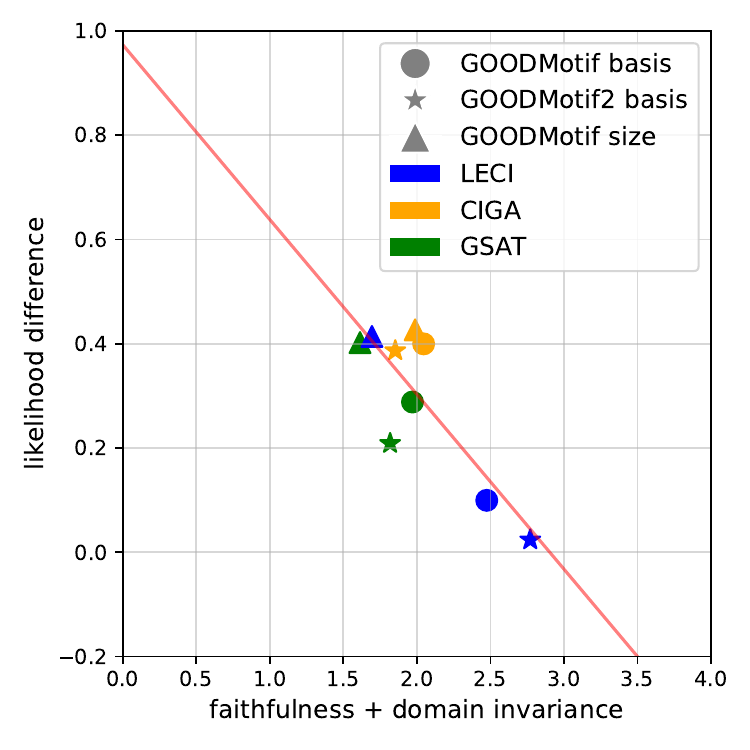}
    \caption{\textbf{Likelihood, faithfulness and domain-invariance are correlated}.  The plot shows the difference in likelihood between splits. The red line is the best linear fit. Best viewed in color.}
    \label{fig:empirical_lowerbound}
\end{wrapfigure}

We stress that while necessity does not appear in the bound, it becomes important if we wish the invariant subgraph to be non-trivial. 
Indeed, \cref{fig:empirical_lowerbound} shows a statistically significant anti-correlation (Pearson's correlation $-0.83$, $p$-value $0.01$) between the difference in average prediction's likelihood between  ID and OOD data, and the combination of the degree of domain-invariance and faithfulness. 
Notably, the same applies, yet with a slightly weaker correlation of $-0.74$ (p-value $0.02$), when replacing likelihood with accuracy. See \cref{app:experimental_details} for further details.


\section{Related Work}
\label{sec:related-work}

\textbf{Modular architectures.}  Modular GNNs have become a popular architectural design choice for seeking trustworthiness, either in the form of explainable \cite{lin2020graph, serra2022learning, miao2022interpretablerandom, kosan2023robust} or confounding-free predictions \cite{wu2022discovering, chen2023rethinking, gui2023joint, chen2022learning} (see \citet{jiang2023survey} for a detailed survey), also in deep neural networks for non-relational data \cite{Koh2020, marconato2022glancenets}.
In both settings, the quality of the explanation -- or any other sparse representation used to communicate across the two modules -- is known to play a crucial role, for users and stakeholders, and for ensuring fair and generalizable model outputs \cite{longa2022explaining, amara2022graphframex, miao2022interpretable}. Despite being designed primarily for faithfulness, as the prediction depends on the explanation only, little attention has been paid to measuring \emph{how well} the classifier adheres its own explanations, an issue that can completely prevent trust allocation \cite{agarwal2024faithfulness}.
Indeed, recent work has shown how for popular \SEGNNs architectures, the classifier seems to be more faithful to randomly generated subgraphs than to its own explanations \cite{christiansen2023faithful}.
In this work, we investigate the reasons for this behavior, discuss the applicability of faithfulness-enforcing strategies and extend the analysis to \DIGNNs.

\textbf{Faithfulness.}  In the Explainable AI \citep{samek2021explaining} and GNN literature \citep{agarwal2023evaluating}, sufficiency and faithfulness are often conflated, however explanations that are not necessary cannot be faithful, as they open to trivially sufficient yet uninformative explanations.
Faithfulness is also sometimes defined as the correlation between the explainers ``performance'' and that of the model, that is, to what extent corrupting the input alters the model's performance \citep{sanchez2020evaluating}. However, this is narrow to model and explainer performance and does not reflect the degree to which a classifier relies on the provided explanation for making its prediction.
Our notion of faithfulness is rooted in notions from causal explainability \citep{beckers2022causal} and disentanglement in causal representation learning \citep{suter2019robustly, scholkopf2021toward}.
Like \citet{beckers2022causal}, we argue that absolute measurements of sufficiency, and more generally faithfulness, are uninformative unless properly contextualized, in our case by the choice of parameters used for the measurements.  Following \citep{bordt2022post}, we argue this leaves room for adversarial explanation providers to supply explanations that optimize their objective rather than that of the explanation consumer.
\correction{}{Alternative techniques for estimating the faithfulness of explanations also exist, either based on data or model randomization \cite{adebayo2018sanity, yonarevisiting}. The former, however, requires manipulations of the underlying data distribution, while the latter is only testing whether the explanation is a function of model parameters, ignoring its impact on models' predictions.}

\section{Discussion and Broader Impact}
\label{sec:conclusion}

We have studied the faithfulness of explanations in regular and modular GNNs.
Our results indicate that, despite conforming to a shared template, existing faithfulness metrics are \textit{not} interchangeable and in fact some even suffer from systematic issues, which we show can be avoided with an appropriate choice of hyperparameters.
\correction{We have shown that optimizing for faithfulness is not always a sensible design goal, and analyzed -- theoretically and empirically -- both strategies for improving faithfulness in modular GNNs and the importance of faithfulness for domain invariance.}{We have also shown that optimizing for faithfulness is not always a sensible design goal, and that improving the faithfulness of \SEGNNs is non-trivial. This highlights limitations in existing models and suggests that better faithfulness could be achieved by improving explanation stability instead, as we plan to do in future work.}
\correction{}{We also proved that faithfulness plays a surprisingly central role for domain invariance, in that modular GNNs tailored for domain invariance cannot even in principle be invariant unless their explanations are also sufficient. This point suggests that research on \DIGNNs should not neglect evaluating sufficiency, and that one could build better \DIGNNs by encouraging the classifier to be faithful.}
From a societal perspective, the aim of this work is to shed light on the non-trivial nature of explanation faithfulness, warning against its unqualified reporting and interpretation. It can  thus contribute to the development of truly interpretable and trustworthy models for networked data.

\correction{}{\textbf{Limitations:} Our experiments focus on four strategies for improving faithfulness rooted in our theory, and show that one of them is \correction{consistently}{particularly} beneficial when applied to SOTA GNN architectures; an extensive evaluation of other approaches, including heuristics (e.g. those listed in the paper), is left to future work.
In addition, our proposed notion of sufficiency hinges on input-level interventions replacing the complement of the explanation subgraph with a different, random complement. Despite not affecting training nor inference time, this operation can become time-consuming for evaluating sufficiency when the input graphs are large, and more efficient implementations should be devised.}



\bibliographystyle{unsrtnat}
\bibliography{main, explanatory-supervision}

\newpage
\section*{NeurIPS Paper Checklist}

\begin{enumerate}

\item {\bf Claims}
    \item[] Question: Do the main claims made in the abstract and introduction accurately reflect the paper's contributions and scope?
    \item[] Answer: \answerYes{} 
    \item[] Justification: The main claims are spelled out in both the abstract and introduction, with remainders to individual sections each providing either theoretical or numerical support for said claims.

\item {\bf Limitations}
    \item[] Question: Does the paper discuss the limitations of the work performed by the authors?
    \item[] Answer: \answerYes{} 
    \item[] Justification: Most of our work is about highlighting issues with existing metrics.  We briefly discuss limitations of our setup in \cref{sec:conclusion}.

\item {\bf Theory Assumptions and Proofs}
    \item[] Question: For each theoretical result, does the paper provide the full set of assumptions and a complete (and correct) proof?
    \item[] Answer: \answerYes{}{} 
    \item[] Justification: Formal proofs for all our theoretical claims are reported in \cref{sec:proofs}.

\item {\bf Experimental Result Reproducibility}
    \item[] Question: Does the paper fully disclose all the information needed to reproduce the main experimental results of the paper to the extent that it affects the main claims and/or conclusions of the paper (regardless of whether the code and data are provided or not)?
    \item[] Answer: \answerYes{} 
    \item[] Justification: Due to space limitations, we could not fully describe our experimental setup in the main text.  However, all details can be found in \cref{app:experimental_details} and in the provided source code.

\item {\bf Open access to data and code}
    \item[] Question: Does the paper provide open access to the data and code, with sufficient instructions to faithfully reproduce the main experimental results, as described in supplemental material?
    \item[] Answer: \answerYes{} 
    \item[] Justification: The source code of our empirical analysis is provided in the supplementary material.  It will also be published on GitHub under an open source license upon acceptance.

\item {\bf Experimental Setting/Details}
    \item[] Question: Does the paper specify all the training and test details (e.g., data splits, hyperparameters, how they were chosen, type of optimizer, etc.) necessary to understand the results?
    \item[] Answer: \answerYes{} 
    \item[] Justification: In \cref{app:experimental_details}.

\item {\bf Experiment Statistical Significance}
    \item[] Question: Does the paper report error bars suitably and correctly defined or other appropriate information about the statistical significance of the experiments?
    \item[] Answer: \answerYes{} 
    \item[] Justification: Our plots and tables include error bars indicating standard deviation for all measures.  For most results, variance information was collected from $5$ runs (varying the seed).  For faithfulness estimation, we report an extended analysis of stability in the appendix.

\item {\bf Experiments Compute Resources}
    \item[] Question: For each experiment, does the paper provide sufficient information on the computer resources (type of compute workers, memory, time of execution) needed to reproduce the experiments?
    \item[] Answer: \answerYes{} 
    \item[] Justification: All experiments were fit and evaluated on a server equipped with A30 and A100 nVidia GPUs.  The models themselves are not computationally expensive to train.
    
\item {\bf Code Of Ethics}
    \item[] Question: Does the research conducted in the paper conform, in every respect, with the NeurIPS Code of Ethics \url{https://neurips.cc/public/EthicsGuidelines}?
    \item[] Answer: \answerYes{} 
    \item[] Justification: All authors read and followed the code of conduct.  Our work does rely on experiments with human subjects or crowd-workers.  All data sets we use are freely available for research purposes and representative of the tasks we address, and they are in fact used in recent related works.  We believe our analysis poses no immediate harmful consequences.

\item {\bf Broader Impacts}
    \item[] Question: Does the paper discuss both potential positive societal impacts and negative societal impacts of the work performed?
    \item[] Answer: \answerYes{} 
    \item[] Justification: Our work is, chiefly, an analysis of faithfulness metrics.  One of our key goals is to identify potential risks of existing metrics, and our results do so for a range of choices, including the one we use in our experiments.  We briefly discuss societal impact in \cref{sec:conclusion}.

\item {\bf Safeguards}
    \item[] Question: Does the paper describe safeguards that have been put in place for responsible release of data or models that have a high risk for misuse (e.g., pretrained language models, image generators, or scraped datasets)?
    \item[] Answer: \answerNA{} 
    \item[] Justification: Our results pose no such risks.

\item {\bf Licenses for existing assets}
    \item[] Question: Are the creators or original owners of assets (e.g., code, data, models), used in the paper, properly credited and are the license and terms of use explicitly mentioned and properly respected?
    \item[] Answer: \answerYes{} 
    \item[] Justification: All data sets and models are properly credited.  All of them can be -- and have already been -- used for research purposes.

\item {\bf New Assets}
    \item[] Question: Are new assets introduced in the paper well documented and is the documentation provided alongside the assets?
    \item[] Answer: \answerYes{} 
    \item[] Justification: The only new asset is the source code for our experiments, which contains minimal instructions for enabling replicability and is available under a non-restrictive license.  The documentation will be improved for the final code release.

\item {\bf Crowdsourcing and Research with Human Subjects}
    \item[] Question: For crowdsourcing experiments and research with human subjects, does the paper include the full text of instructions given to participants and screenshots, if applicable, as well as details about compensation (if any)? 
    \item[] Answer: \answerNA{} 
    \item[] Justification: The paper does not involve crowdsourcing nor research with human subjects.

\item {\bf Institutional Review Board (IRB) Approvals or Equivalent for Research with Human Subjects}
    \item[] Question: Does the paper describe potential risks incurred by study participants, whether such risks were disclosed to the subjects, and whether Institutional Review Board (IRB) approvals (or an equivalent approval/review based on the requirements of your country or institution) were obtained?
    \item[] Answer: \answerNA{} 
    \item[] Justification: The paper does not involve crowdsourcing nor research with human subjects.

\end{enumerate}

\appendix
\newpage

\section{Proofs}
\label{sec:proofs}

\subsection{Proof of \cref{prop:metrics-are-not-equivalent}}

\propmetricsnotequiv*

\begin{proof}
We prove the statement for graph classification; the case of node classification is analogous.

Assume $\RX$ attains a perfect degree of sufficiency according to $\PR$, \ie $\SUF_{\PR}(\RX) = 0$, and take $\PR'$ such that $\calS \cap \calS' = \emptyset$, where we shortened $\calS = \mathrm{supp}(\PR)$ and $\calS' = \mathrm{supp}(\PR')$.  Also, assume that both supports are finite.
Then:
\[
    |\SUF_{\PR}(\RX) - \SUF_{\PR'}(\RX)|
        = \Big|
            \sum_{\GX' \in \calS} \Delta(\GX, \GX') \PR(\GX')
            + \sum_{\GX' \in \calS'} \Delta(\GX, \GX') \PR'(\GX')
        \Big|
\]
The two sums have no terms in common, so we can independently control the second one through our choice of interventional distribution $\PR'$.  In fact, it is maximized by those distributions that associate all probability mass to graphs $\GX'$ for which $\Delta(\GX, \GX')$ is largest, \eg counterfactuals.
As long as at least one counterfactual $\GX'$ exists that includes $\RX$ as a subgraph, then $\PR'$ can allocate all probability mass to it, in which case the right sum becomes as large as $\argmax_{\vp, \vq} d(\vp, \vq)$, that is, as large as the natural range of the divergence $d$.

A similar reasoning applies to $\NEC$.  Take $\RX$ such that for all $\RX' \in \supp(\PC)$ it holds that $\Delta(\GX, \GX') > \tau$ for some desired threshold $\tau$.  As long as there exists a graph \correction{$\GX' \supseteq \CX$}{$\GX' \supseteq \RX$} such that $p_\theta(\cdot \mid \GX) \equiv p_\theta(\cdot \mid \GX')$, then we can always choose an interventional distribution \correction{$\PC$}{$\PC'$} that allocates all mass to it.  Hence, $\NEC_{\PC'}(\RX) = 0$ and the difference in necessity will be at least $\tau$.
\SA{SA-ST: Check me}
\end{proof}

This simple result warrants some discussion.
First, we note that the assumptions it hinges on are rather weak.  Its two key requirements are that:
i) $\calS$ and $\calS'$ are finite: this is the case for all existing metrics, whose interventional distributions are defined over the set of \textit{subgraphs} of the input graph, which are finitely many;
ii) The GNN admits counterfactuals that subsume $\RX$ (for sufficiency) or inputs that subsume $\CX$ and map to the predicted label $\hat y$ (for necessity), both frequent occurrences in practice.

Second, the construction behind the proof also mimics actual differences between faithfulness metrics from the literature.
In fact, ignoring differences in choice of divergence, the main difference between unfaithfulness (which allocates non-zero probability only to subgraphs obtained by deleting features from the input), fidelity minus (which deletes all edges), and robust fidelity minus (which deletes a random subset of edges at random) is exactly their interventional distributions $\PR$, which also happen to have either disjoint or almost disjoint supports (in which case \correction{the}{} a similar result applies almost verbatim).
Our construction then amounts to saying that there are many practical situations in which one can achieve good unfaithfulness and poor fidelity minus, or vice versa.  

Let us briefly discuss the relationship between metrics having the same interventional distribution but different divergences $d$ and $d'$.
On the bright side, it is easy to see that, as long as both $d$ and $d'$ are proper divergences, an explanation $\RX$ achieving perfect sufficiency according to one will also attain perfect sufficiency according to the other, precisely because perfect sufficiency is attained when $\Delta(\GX, \GX') = 0$ for all $\GX' \sim \PR$, which can occur if and only if both divergenes are zero (by definition of divergence).  The same holds for the degree of necessity.
For non-optimal sufficiency and necessity, the difference due to replacing divergences is governed by well-known inequalities, such as H\"older's (between $L_p$ distances) and Pinsker's (relating the \KL to the $L_1$ distance).

The situation is different if $d$ is a proper divergence (say, the \KL divergence) and $d'$ is a difference in likelihoods.  In this case, there are situations in which the two quantities can differ.
To see this, consider a multi-class classification problem, target decision $(\GX, \hat y)$ and a modified input $\GX' \sim \PR$ such that i) $p_\theta(\hat y \mid \GX) = p_\theta(\hat y \mid \GX')$, but ii) the two label distributions have disjoint support.  In this case, the \KL divergence between them would be unbounded, yet the difference in likelihood would be null.
Depending on the choice of $\PR$, this can yield a large difference between degrees of sufficiency (and similarly for necessity).

\subsection{Proof of \cref{prop:rfid_expval}}

\proprfidexpval*

\begin{proof}
    \RFIDP corrupts the explanation $R$ by deleting edges independently at random. Formally, let $U_j$ be a random variable determining whether edge $e_j$ is kept, such that $U_j \sim Bernoulli(\kappa)$, with $\kappa$ a user-provided hyperparameter.
    As in \cref{prop:nec_expval}, let $\calS_R$ be the set of all subgraphs of $G$ obtained by leaving the complement of $R$ unmodified, and $\calA_R$ those subgraphs that lead to a large enough change in likelihood, that is, $\calA_R = \{ G' \in \calS_R : \Delta(G, G') \ge \epsilon \}$, we can study \RFIDP in terms of how likely the corrupted graphs $G'$ it samples have missing relevant edges, \ie $P(G' \in \calA_R)$.
    \begin{align}
        P(G' \in \calA_R)
            & = 1 - P(G' \notin \calA_R)\\
            & = 1 - P(U_1=1, \ldots, U_m=1)\\
            & = 1 - \prod_{j=1}^m P(U_j=1)\\
            & = 1 - \kappa^m\\
    \end{align}
    In the second to last step, we made use of the independence of all $U_i$'s. 
    \correction{which does not depend on the number of irrelevant edges $\norm{R} - r$}{The above probability does not depend on the number of irrelevant edges $\norm{R} - r$}.
\end{proof}

\subsection{Proof of \cref{prop:nec_expval}}

\propnecexpval*

\begin{proof}
    Let \PC be the uniform distribution over $\calS_R$.  Then, the probability of sampling a perturbed graph $G' \sim \PC(G)$ such that $G' \in \calA_R$ is exactly $|\calA_R|/|\calS_R|$.
    Now, $|\calA_R|$ is exactly $(2^r - 1) \cdot 2^{\norm{R} - r}$, while $|\calS_R|$ is $2^{\norm{R}} - 1$.  Hence, \NEC depends on both the number of truly relevant edges $r$ and on the size of the explanation, as expected.
    When choosing instead a budget \budget of deletions, and a uniform $\PC(G)$ over subgraphs with $\norm{G} - \budget$ edges, $P(G' \in \calA_R^\budget) = |\calA_R^\budget|/|\calS_R^\budget|$.
    Note that $|\calS_R^\budget| = \binom{\norm{R}}{\budget}$, while $|\calA_R^\budget| = \sum_{c = 1}^\budget \binom{r}{c} \binom{\norm{R} - r}{r - c}$.
\end{proof}

\subsection{Proof of \cref{prop:sse-are-trivial}}

\propssearetrivial*

\begin{proof}
    Before proceeding, it is useful to introduce the notion of computational graph, that is, the subgraph of $\GX$ induced by those nodes whose messages are relevant for determining the label distribution $P(Y_u \mid \GX)$ or $P(Y \mid \GX)$.
    For node classification, it is easy to see that for GNNs with local aggregators, the only messages reaching $u$ are those coming from nodes within $N_L(u)$, and for injective GNNs, all such messages impact the label distribution.
    For graph classification, which is implicitly global, all nodes in the computational graph are the entirety of $\GX$.

    Next, we show that if $\RX$ does not subsume the entirety of the computational graph, then when computing strict sufficiency we can always alter those elements of the computation graph that fall in the complement $\CX$.
    By injectivity of $p_\theta$, any change to the features or edges of the computational graph yields a difference in label distribution, necessarily affecting both classes and therefore any divergence or difference in likelihood $d$.
    It follows that $\RX$ cannot be strictly sufficient for any $d$, and therefore neither strictly faithful.

    The converse also holds: if $\RX$ covers the entire computational graph, then it is necessarily strictly sufficient.  This is because the complement $\CX$ has no overlap with the computational graph, so altering it has no impact on the label distribution.
\end{proof}

\subsection{Proof of \cref{prop:suff_dignns}}

\suffdignns*

\begin{proof}
    We proceed by contradiction. Let $\GX$ be an input graph such that the explanation $\RX$ for model $\MONO$ and decision $(\GX, \hat{y})$ is perfectly domain-invariant 
    yet not strictly sufficient. We assume that the model prediction does not depend on domain-induced spurious information, meaning that every modification applied to $\CX$ will not have any impact on the model prediction (according to $d$). 
    However, the lack of strict sufficiency implies that $\exists \GX' \in \mathrm{supp}(\PR)$ such that $\Delta(\GX, {\GX}') > 0$. 
    This means that there exists a perturbation outside of $\RX$ that altered the model prediction, which is in contradiction with assuming that it does not depend on domain-induced information.
\end{proof}

\subsection{Proof of \cref{thm:faith_ood_upper_bound}}

\textbf{Theorem:} \textit{
Let $\MONO$ be a deterministic \DIGNN with detector \DET and classifier \CLF and $d$ be the difference in likelihood of the predicted label.
Also, let $\MONO$ be Lipshitz with respect to changes to both the topology and the features of the input, that is, for every pair of graphs $\GX = ( G, X )$ and $\GX' = ( G', X' )$, it must hold that:
\begin{align}
    \bigg | \MONO (Y \mid ( G, X )) - \MONO(Y \mid ( G', X )) \bigg |
        & \le k_1 d_1(G, G')
    \\
    \bigg | \MONO(Y \mid ( G, X )) - \MONO(Y \mid ( G, X' )) \bigg |
        & \le k_2 d_2(X, X')
\end{align}
for suitable distance functions $d_1$ and $d_2$.
%
%
Then:
\begin{align}
    \label{eq:faith_ood_upper_bound}
        \Big | \expectedGy{p^{id}} [\MONO(y \mid \GX)] \ - \ & \expectedGy{p^{ood}} [\MONO(y \mid \GX)] \Big | 
    \\
    & \le 
        \bbE_{\GXInv} [
            k_1 (\LAMBDATOPO^{id}+\LAMBDATOPO^{ood}) + k_2 (\LAMBDAFEAT^{id}+\LAMBDAFEAT^{ood}) + (\LAMBDASUFF^{id} + \LAMBDASUFF^{ood}) \nonumber
        ]
\end{align}
Here, $k_1, k_2 > 0$ are Lipschitz constants of \MONO, $\LAMBDATOPO^{id}$, $\LAMBDATOPO^{id}$ $\LAMBDAFEAT^{ood}$, and $\LAMBDAFEAT^{ood}$ measure the (negated) degree of domain invariance of the detector's output (with respect to topology and features, respectively), and $\LAMBDASUFF^{id}, \LAMBDASUFF^{ood}$ are the degree of sufficiency for ID and OOD data.
}

\begin{proof}
Let $\DET^*$ be an \emph{ideal} detector that for every input \GX outputs the maximal domain-invariant subgraph $\RX^* = ( R^*, X^{R^*} )$.
Note that, for every $(\GX^{id}, y^{id}) \sim p^{id}$ and $(\GX^{ood}, y^{ood}) \sim p^{ood}$ that share the same invariant subgraph (that is, such that $f^*(\GX^{id}) = f^*(\GX^{ood}) = \GXInv$) any \DIGNN constructed by stitching together $f^*$ and a deterministic classifier \CLF will predict the same label distribution for both graphs, \ie $p_{\CLF}(Y \mid f^*(\GX^{id})) \equiv p_{\CLF}(Y \mid f^*(\GX^{ood}))$.

Now, fix any invariant subgraph $\GXInv$.
When conditioning on it, it holds:
\begin{align}
    \expectedCondGy{p^{id}}{\GXInv} [ \MONO(y \mid \DET^*(\GX^{id})) ]
    = \expectedCondGy{p^{ood}}{\GXInv} [ \MONO(y \mid \DET^*(\GX^{ood})) ]
\end{align}
where the two expectations run over ID and OOD samples that share the same invariant subgraph $\GXInv$.
We proceed by noting that:
\begin{align}
    \expectedCondInvGy{p^{id}} \bigg |  \MONO(y \mid (R^*, X^{R^*})) - \MONO(y \mid (R, X^{R^*})) \bigg |
        & \leq k_1 \underbrace{ \expectedCondInvGy{p^{id}} [ d_1(R^*, R) ] }_{:= \LAMBDATOPO}
        \label{eq:div_plaus}
\end{align}
The expectation on the \textcolor{red}{RHS} is the average topological distance of the predicted invariant subgraphs to $\GXInv$, \correction{}{which corresponds to the degree of invariance (plausibility) of $R$,} and will be referred to as $\LAMBDATOPO$.
At the same time, it also holds that:
\begin{align}
    \expectedCondInvGy{p^{id}} \bigg |  \MONO(y \mid (R, X^{R^*})) -  \MONO(y \mid (R, X^{R})) \bigg |
        & \leq k_2 \underbrace{ \expectedCondInvGy{p^{id}} [ d_2(X^{R^*} , X^{R}) ] }_{:= \LAMBDAFEAT}
        \label{eq:div_feat_plaus}
\end{align}
where the expectation on the RHS is now the average distance between the features of the predicted invariant subgraphs and those of $\GXInv$, and will be referred to as $\LAMBDAFEAT$.
Combining \cref{eq:div_plaus} and \cref{eq:div_feat_plaus} using the triangular inequality, yields: \SA{Made application of triangular inequality explicit}
%
{
\color{red}
\begin{align}
\label{eq:div_bound_id}
&
        \expectedCondInvGy{p^{id}} \bigg | 
            \MONO(y \mid \RX^*) -  \MONO(y \mid \RX) 
        \bigg | 
    \\
& = 
        \expectedCondInvGy{p^{id}} \bigg | 
            \MONO(y \mid \RX^*) - \MONO(y \mid (R, X^{R^*})) + \MONO(y \mid (R, X^{R^*})) - \MONO(y \mid \RX) 
        \bigg | 
    \\
& \leq
            \bbE \bigg |  \MONO(y \mid (R^*, X^{R^*})) - \MONO(y \mid (R, X^{R^*})) \bigg |
        +
            \bbE \bigg |  \MONO(y \mid (R, X^{R^*})) -  \MONO(y \mid (R, X^{R})) \bigg |
    \\
& \leq 
        k_1 \LAMBDATOPO^{id} + k_2 \LAMBDAFEAT^{id}
\end{align}
}
%
%
This holds for any fixed $\GXInv$.

Next, we bound the expected difference between the label distribution determined by $\RX$ and that determined by $\GX$ using the degree of sufficiency.
For notational convenience, we draw complements $\CX'$ rather than full modified graphs $\GX'$ from $\PR$, and we denote the operation of joining $\RX$ and $\CX'$ to form $\GX'$ with the $\cup$ operator, and where we use the difference in prediction likelihood as divergence $d$.
We proceed as follows:
\begin{align}
\label{eq:div_suff}
    &
        \expectedCondInvGy{p^{id}} 
            \bigg | 
               \MONO(y \mid \RX)  -  \MONO(y \mid \GX) 
            \bigg |
    \\
    & =
        \expectedCondInvGy{p^{id}} 
            \bigg | 
                \bbE_{\CX' \sim \PR(\GX)} 
                    \big [ \MONO(y \mid \RX \cup \CX') \big ] -  \MONO(y \mid \GX) 
            \bigg | 
    \\
    & \le
        \expectedCondInvGy{p^{id}}
            \bbE_{\CX' \sim \PR(\GX)} 
            \bigg | 
                \MONO(y \mid \RX \cup \CX') -  \MONO(y \mid \GX) 
            \bigg | 
    \\
    & = \expectedCondInvGy{p^{id}} 
            \SUF(\RX) \defeq \LAMBDASUFF \\
\end{align}
%
%
In the first to second step, we made use of the law of total probability and the product rule, for which:
\begin{align}
    \MONO(y \mid \RX)
        & = \sum_{\CX'} p(y, \CX' \mid \RX) 
        = \sum_{\CX'} \frac{p(y, \CX', \RX)}{p(\RX)}
        = \sum_{\CX'} \frac{p(y \mid \CX', \RX)p(\CX' \mid \RX)p(\RX)}{p(\RX)}
    \\
        & = \sum_{\CX'} \MONO(y \mid \RX \cup \CX') \PR(\CX')
        = \bbE_{\CX' \sim \PR} \MONO(y \mid \RX \cup \CX')
\end{align}

Then, applying again the triangular inequality between \cref{eq:div_suff} and \cref{eq:div_bound_id} yields:
\begin{equation}
\label{eq:div_bound_id_final}
    \expectedCondInvGy{p^{id}} \bigg | \MONO(y \mid \RX^*) -  \MONO(y \mid \GX) \bigg | \leq k_1 \LAMBDATOPO^{id} + k_2 \LAMBDAFEAT^{id} + \LAMBDASUFF^{id}.
\end{equation}
The same derivation applies to $p^{ood}(\GX)$, so we obtain:
\begin{equation}
\label{eq:div_bound_ood}
    \expectedCondInvGy{p^{ood}} \bigg | \MONO(y \mid \RX^*) -  \MONO(y \mid \GX) \bigg | \leq k_1 \LAMBDATOPO^{ood} + k_2 \LAMBDAFEAT^{ood} +  \LAMBDASUFF^{ood},
\end{equation}
Combining these two bounds using the triangular inequality one last time, we derive: \SA{Made application of Triangular Inequality explicit}
{
\color{red}
\begin{align}
\label{eq:div_bound_id_ood}
&
        \bigg | \expectedCondInvGy{p^{id}} \MONO(y \mid \GX) -
         \expectedCondInvGy{p^{ood}} \MONO(y \mid \GX) \bigg |
    \\
& \leq 
        \bbE_{\substack{(\GX^{id}, y) \sim p^{id}(\cdot \mid \GXInv) \\ (\GX^{ood}, y) \sim p^{ood}(\cdot \mid \GXInv)}} 
            \bigg | \MONO(y \mid \GX^{id}) -
                    \MONO(y \mid \GX^{ood}) 
            \bigg |
    \\
& \leq 
        k_1 (\LAMBDATOPO^{id}+\LAMBDATOPO^{ood}) + k_2 (\LAMBDAFEAT^{id}+\LAMBDAFEAT^{ood}) + (\LAMBDASUFF^{id} + \LAMBDASUFF^{ood}).
\end{align}
}
Again, this holds for all choices of $\GXInv$.

Finally, we leverage these inequalities to bound the difference in likelihood between ID and OOD data \correction{}{for all possible choices of $\GXInv$}:
\begin{align}
    & \bbE_{\GXInv} \left[ \left|
            \expectedGy{p^{id}(\cdot \mid \GXInv)}  \MONO(y \mid \GX) -
            \expectedGy{p^{ood}(\cdot \mid \GXInv)} \MONO(y \mid \GX)
        \right| \right] 
    \\
    & \geq
        \bigg | \bbE_{\GXInv} \left[ \expectedGy{p^{id}(\cdot \mid \GXInv)}  \MONO(y \mid \GX) -
                 \expectedGy{p^{ood}(\cdot \mid \GXInv)} \MONO(y \mid \GX) \right] \bigg |
    \\
    & =
        \bigg | \expectedGy{p^{id}}  \MONO(y \mid \GX) -
                 \expectedGy{p^{ood}} \MONO(y \mid \GX) \bigg |
\end{align}
where $\bbE_{\GXInv}$ runs over all possible maximally invariant subgraphs and no longer depends on the specific choice of 
$\GXInv$.
By monotonicity of the expectation we conclude that:
\begin{align}
    \bigg | \expectedGy{p^{id}}  \MONO(y \mid \GX) - &
             \expectedGy{p^{ood}} \MONO(y \mid \GX) \bigg |
    \\
    & \leq \bbE_{\GXInv} \left[ k_1 (\LAMBDATOPO^{id}+\LAMBDATOPO^{ood}) + k_2 (\LAMBDAFEAT^{id}+\LAMBDAFEAT^{ood}) + (\LAMBDASUFF^{id}+\LAMBDASUFF^{ood}) \right]
\end{align}

\correction{}{which permits a practical estimation of the RHS by averaging over a set of samples without the need to condition explicitly on a specific invariant subgraph.}

\end{proof}

\section{Experimental Details}
\label{app:experimental_details}

\subsection{Datasets}\label{app:dataset}
In this study, we conduct an investigation across seven graph classification datasets commonly used for evaluating \SEGNNs and \DIGNNs. Specifically, we examine three synthetic datasets and four real-world datasets, and in the following paragraphs, we provide a detailed description of each.
\paragraph{Synthetic datasets}
\begin{itemize}[leftmargin=1.25em]
    \item \BAMULTISHAPES \cite{azzolin2022global} is a synthetic dataset consisting of 1,000 Barabasi-Albert (BA) graphs, each with network motifs (house, grid, wheel) randomly attached at various positions. Class 0 includes plain BA graphs and BA graphs enriched with either a house, a grid, a wheel, or all three motifs together. Class 1 consists of BA graphs enriched with a combination of two motifs: a house and a grid, a house and a wheel, or a wheel and a grid. This dataset is utilized in the context of self-explainable Graph Neural Networks (GNNs), with the expected ground truth explanations being the specific motif combinations in each class.
    
    \item \MotifTwoBasis \cite{gui2023joint} is a synthetic dataset comprising 24,000 graphs categorized into three classes. Each graph consists of a basis and a motif. The basis can be a ladder, a tree (or a path), or a wheel. The motifs are a house (class 0), a five-node cycle (class 1), or a crane (class 2). The dataset is divided into training (18,000 graphs), validation (3,000 graphs), and test sets (3,000 graphs).
    In the context of OOD analysis, two additional sets are considered: the OOD validation set and the OOD test set. In these sets, the class-discriminative subgraph $\RX$ (i.e., the motifs) remains fixed, while the bases \correction{are varied}{vary}. Specifically, the basis for the OOD validation set (3,000 graphs) is a circular ladder, and the basis for the OOD test(3,000 graphs) set is a Dorogovtsev-Mendes graph \cite{dorogovtsev2002evolution}. 

    \item \MotifBasis \cite{gui2022good} is specular to \MotifTwoBasis, where the OOD test set contains graphs generated connecting the three motifs above to simple line paths of varying length.
    
    \item  \MotifSize \cite{gui2022good} is a synthetic dataset consisting of 24,000 graphs categorized into three classes. Similar to \MotifTwoBasis, each network is composed of a basis and a motif, with the motif serving as the class-discriminative subgraph $\RX$. The basis structures include a ladder, a tree (or a path), a wheel, a circular ladder, or a star. The motifs are a house (class 0), a five-node cycle (class 1), or a crane (class 2). The dataset is divided into training (18,000 graphs), validation (3,000 graphs), and test sets (3,000 graphs). 
    In the context of OOD analysis, the size of the basis is increased. Specifically, in the OOD validation set (3,000 networks), the basis sizes are increased up to three times their original size, while in the OOD test set (3,000 \correction{networks}{graphs}), the basis sizes are increased up to seven times.
\end{itemize}

\paragraph{Real-world datasets}
\begin{itemize}[leftmargin=1.25em]
    \item \BBBP \cite{wu2018moleculenet} is a dataset derived from a study on modeling and predicting barrier permeability \cite{martins2012bayesian}. It comprises 2,050 compounds, with 483 labeled as positive and 1,567 as negative. 
    \item \CMNISTS \cite{gui2022good} contains 70,000 graphs of hand-written digits transformed from the MNIST database using superpixel techniques\cite{monti2017geometric}. The digits are colored according to their domains and concepts. In the training set, which contains 50,000 graphs, the digits are colored using five different colors. To evaluate the model's performance on out-of-distribution data, the validation and testing sets each contain 10,000 graphs with two new colors introduced specifically for these sets.
    \item \LBAPS \cite{gui2023joint} is a molecular dataset consisting of 34,179 graphs sourced from the 311 largest chemical assays. The validation set comprises 19,028 graphs from the next largest 314 assays, while the test set includes 19,302 graphs from the smallest 314 assays.    
    \item \SSTTwoS is a sentiment analysis dataset based on the NLP task of sentiment analysis, adapted from the work of Yuan et al. \cite{yuan2022explainability}. In this dataset, each sentence is transformed into a grammar tree graph, where individual words serve as nodes and their associated word embeddings act as node features. The primary task is a binary classification to predict the sentiment polarity of each sentence.The dataset comprises 70,042 graphs, divided into training, validation, and test sets. The out-of-distribution (OOD) validation and test sets are specifically created to evaluate performance on data with longer sentence lengths.
\end{itemize}

\subsection{Implementation Details}
\label{app:impl_details}
\paragraph{Training and reproducibility}
The models are developed leveraging repositories provided by previous work. Specifically:
\begin{itemize}[leftmargin=1.25em]
    \item \CIGA, \LECI and \GSAT are developed based on the repository from \cite{gui2023joint}, using commit \texttt{fb39550453b4160527f0dcf11da63de43a276ad5}.
    \item \RAGE is implemented using the code available at \texttt{https://anonymous.4open.science/r/RAGE} as described in \cite{kosan2023robust}.
    \item For \GISST, we use the repository from \texttt{https://anonymous.4open.science/r/SEGNNEval}, following the approach outlined in \cite{lin2020graph}.
\end{itemize}
To encourage reproducibility, we stick to the hyperparameters provided in each respective repository \correction{}{, except for \GSAT on \BBBP and \BAMULTISHAPES where we set the values of \texttt{ood\_param} to $0.5$ and \texttt{extra\_param} to $[\texttt{True}, 10, 0.2]$, and for \MotifTwoBasis and \MotifSize in \cref{tab:mitigations_SE} where we set the values of \texttt{ood\_param} to $10$ and \texttt{extra\_param} to $[\texttt{True}, 10, 0.2]$.} \SA{Antonio check your hyper-params} Model selection was performed on the ID validation set.


\paragraph{Computing faithfulness}

Our practical implementation of \NEC follows the guidelines presented in \cref{sec:estimators_not_equally_good}. The idea is that choosing a fixed budget depending, for example, on a dataset-wide statistic is a simple yet sensible choice for a necessity metric properly accounting for the number of irrelevant edges present in an explanation. 
Specifically, we chose the budget \budget as a fixed proportion of the average number of undirected edges for each split of the dataset, where the proportion ratio is set to $5\%$ in all our experiments. For each explanation, we sample a number \budgetMC of intervened graphs, where \budgetMC is fixed at $8$ in our experiments.

For \SUF, instead, we replace the complement of the explanation with that of a random sample from the same data split, joining with random connections the original explanation with the sampled complement. We repeat this operation for a number \budgetMC of samples for each explanation, where \budgetMC is fixed at $8$ in our experiments, as for \NEC.
\correction{}{Since the joining operation is performed explicitly at the input level, the resulting procedure can become computationally heavy, especially for large graphs. 
Alternative approaches try to emulate input interventions with latent interventions, where manipulations to the embedding of explanation and complement are applied, respectively \cite{wu2022discovering, Sui2022cal}. 
We argue that this type of latent intervention are suboptimal for a correct evaluation of faithfulness, as the latent vectors used for interventions are not guaranteed to sufficiently disentangle explanation and complement information, resulting in leakages. }

\begin{figure}
    \centering
    \subfigure{\includegraphics[width=1\textwidth]{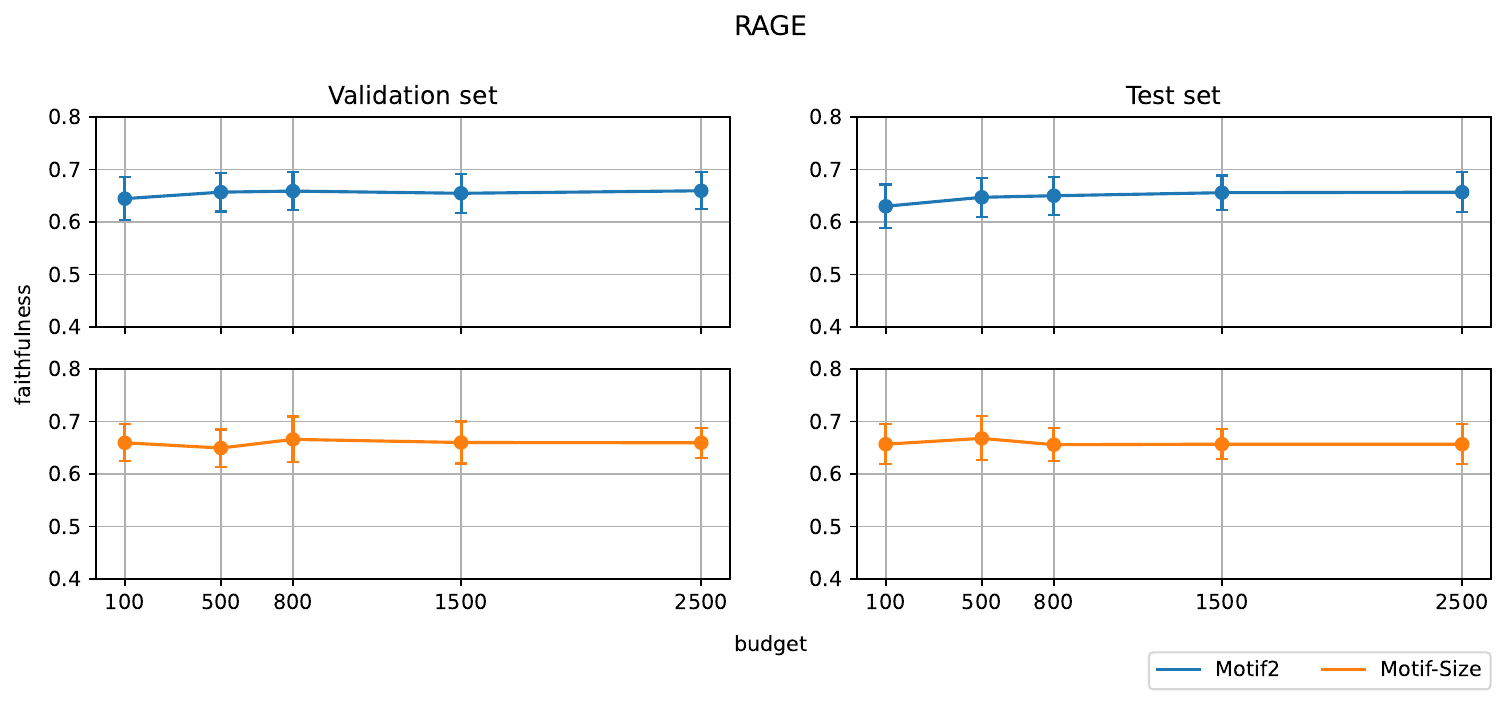}}
    \subfigure{\includegraphics[width=1\textwidth]{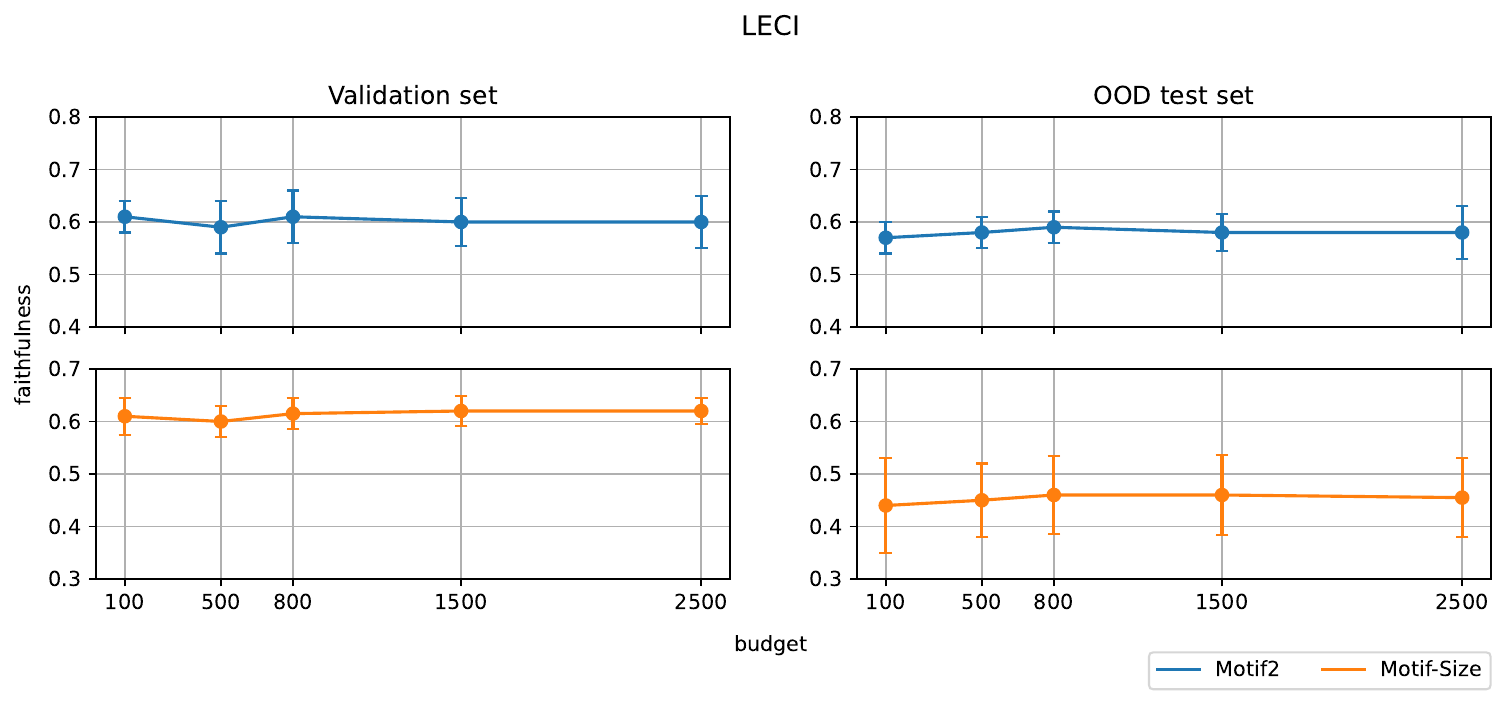}}
    \caption{\textbf{Ablation study of faithfulness scores} across varying number of number of samples \budgetSamples.}
    \label{fig:ablation_numsaplaes}
\end{figure}

\begin{figure}
     \centering
     \subfigure{\includegraphics[width=1\textwidth]{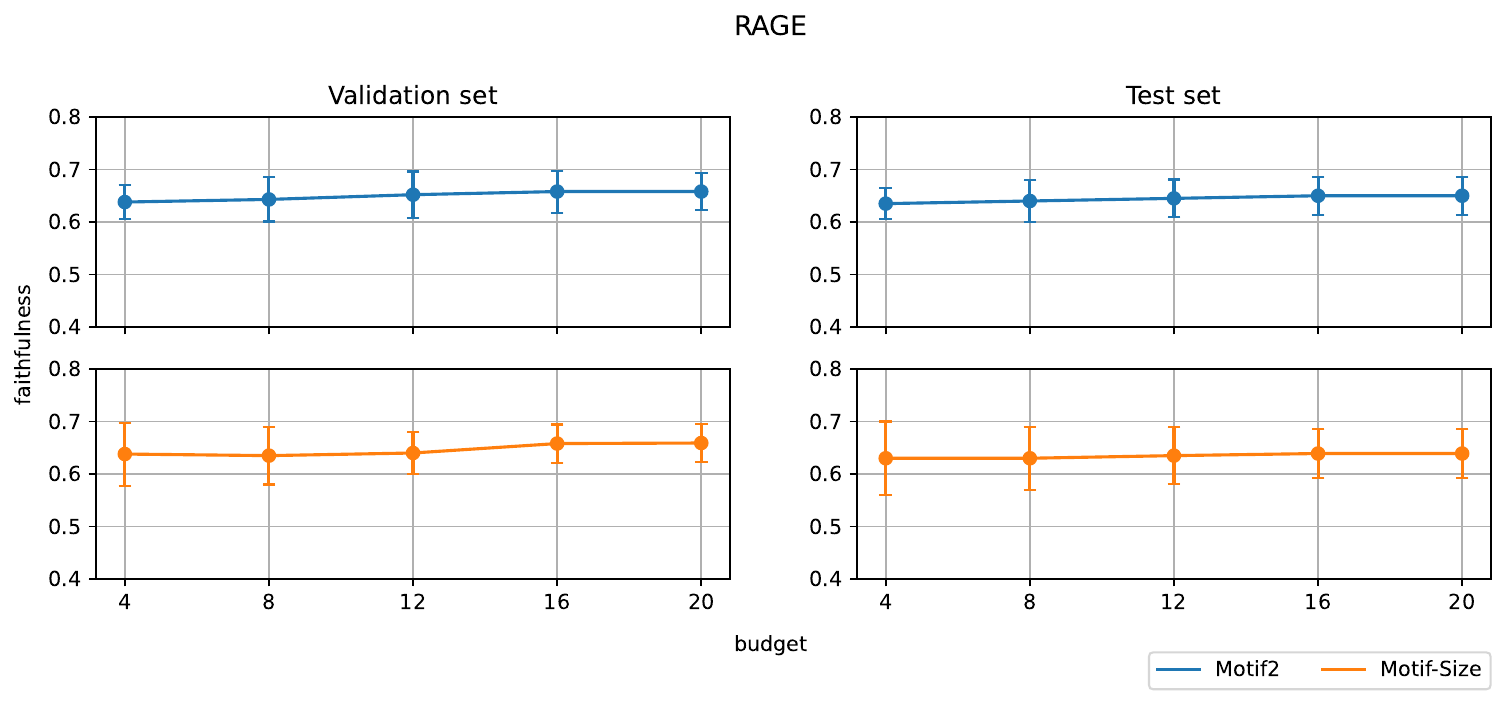}}
     \subfigure{\includegraphics[width=1\textwidth]{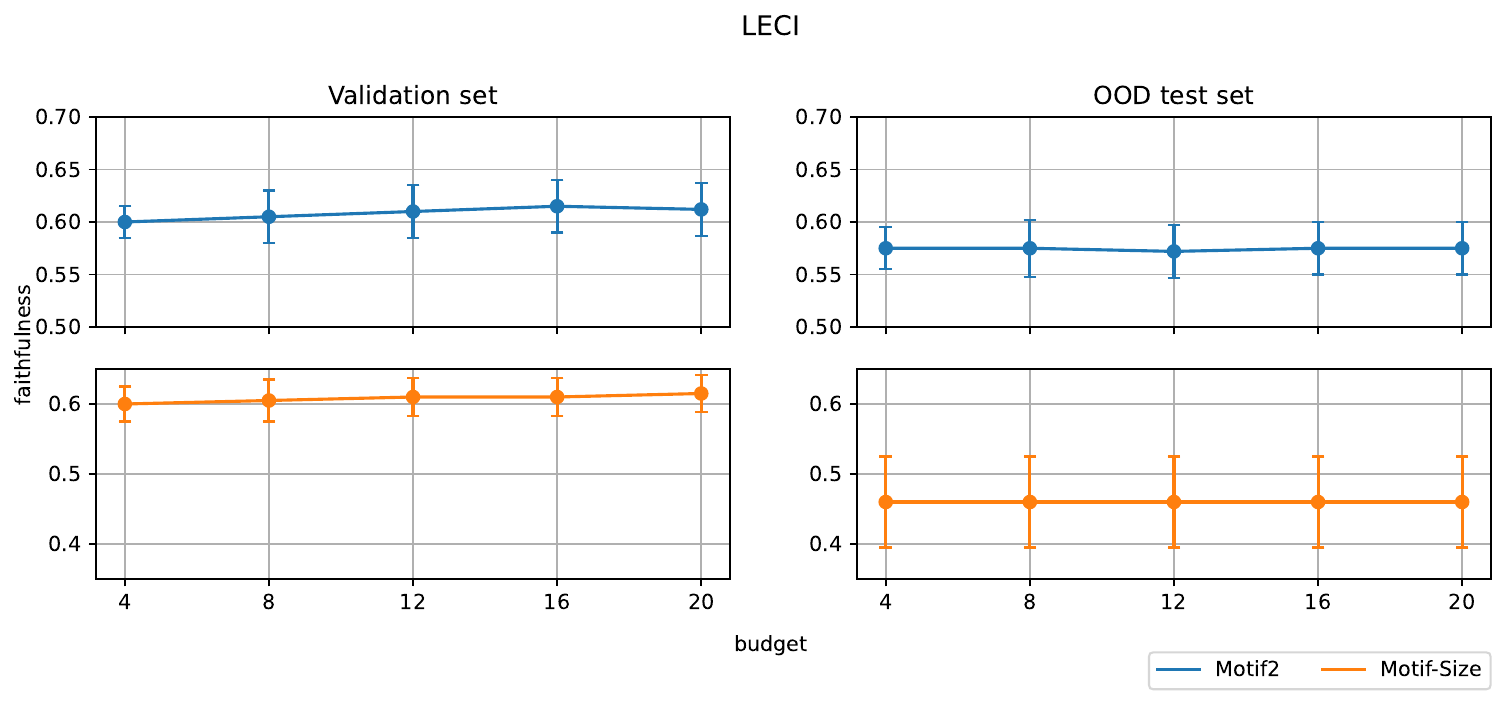}}
     \caption{\textbf{Ablation study of faithfulness scores} across varying number of interventions \budgetMC.}
     \label{fig:ablation_expval}
 \end{figure}

The final faithfulness corresponds to the harmonic mean of the normalized \NEC and normalized \SUF scores, where we set $d$ as the $L_1$ divergence for both metrics, and it is computed over a subset of \budgetSamples of input graphs, which is set to $800$ in our experiments. An ablation study to compare the resulting scores across a varying number of samples \budgetSamples and interventions \budgetMC is provided in \cref{fig:ablation_numsaplaes} and \cref{fig:ablation_expval}.
\correction{}{In all experiments, we use the normalized \SUF and \NEC, such that the values are the higher the better.}

\correction{}{A question to raise is: \emph{are faithfulness metrics well calibrated?} The answer is no, in the sense that a value of, e.g., $0.5$ does not have the same meaning across metrics. This issue affects all existing metrics and is naturally present also in \SUF and \NEC.}

\paragraph{Computing the degree of invariance} The degree of invariance appearing in \cref{thm:faith_ood_upper_bound} represents how close the predicted subgraph $\RX$ is to the truly domain-invariant input motif, which is assumed to be the only stable factor enabling for stable predictions across domains. Carrying over the terminology from XAI, we compute this quantity as the \emph{plausibility} \cite{longa2022explaining} of the provided subgraph with respect to a known ground truth. Specifically, we compute the topology-wise degree of invariance as the Weighted Intersection over Union (WIoU) between the provided explanation relevance score, and the expected ground truth scores, which equal $1$ for the invariant edges in the graph, and $0$ for the others. The feature-wise degree of invariance, instead, is unmeasured as the ground truth over invariant features is typically overlooked, and no settled evaluation testbed is available.

\paragraph{Computing instability of explanations}
\correction{}{  We measure how stable explanations are to input perturbations as the Matthews Correlation Coefficient (MCC) between the clean binarized edge explanation mask and the binarized explanation after perturbing the original explanation complement, averaged over $\{ 30\%, 60\% \}$ top-K binarization thresholds. 
Perturbations are limited to edge removals, and the number of perturbations is a fixed proportion of the average number of edges in the split under evaluation. In our experiments, the proportion is set to $5 \%$ unless otherwise specified, results are computed on the in-distribution validation set, and for each input graph we sample $5$ random perturbations.
Notice that measuring the drop in plausibility (as described in the paragraph above) does not reliably measure how much the explanation changes, since removing elements in the complement makes the denominator of WIoU smaller, thus perturbations tend to artificially increase the score.
For ground truth annotated datasets, perturbations are applied such as not modifying also the ground truth explanations.
We provide in \cref{fig:example_stability} a visual example of the instability of \DIGNNs, where a single edge corruption can drastically change the explanation.
}

\correction{}{  To the best of our knowledge, there is no previous work explicitly studying the stability of \SEGNNs and \DIGNNs, but our findings strongly correlate with those studying the stability of explanations for post-hoc explainers, which highlight severe instabilities to both adversarial and non-adversarial perturbations \cite{longa2022explaining, fang2023evaluating, bajaj2021robust, li2024underfire, wang2023vinfor}. 
We believe this is not a surprise, as the way the detector is typically implemented in \SEGNNs and \DIGNNs is basically a train-time adaptation of the PGExplainer post-hoc explainer \cite{luo2020parameterized}.
%
}

\correction{}{We acknowledge that more advanced techniques for corrupting graphs are possible, possibly including feature-wise interventions. We leave this investigation to future developments.}

\paragraph{Implementation of faithfulness-enforcing strategies}

As stated in Section \ref{sec:perfect_faith_modular}, \correction{we modified the official source code of several models to implement the strategies proposed in Proposition 6}{we included the aforementioned architectural desiderata in several popular models to verify their impact in model accuracy and faithfulness.}

\correction{}{The first important remark is that the benchmarked models were originally customized around a specific architecture, therefore we might expect that some desiderata turn out to be detrimental to model performance when keeping the original hyper-parameters, especially if those were obtained after crafted hyper-parameter optimization. 
On the other hand, heavily changing the hyper-parameters would result in mixed effects, which make the assessment of the four desiderata less reliable. }

\correction{}{The second remark is about the learnability of the proposed desiderata. 
In fact, they represent ideal behaviors the detector and the classifier should have, but this does not necessarily translate into easy-to-learn models. For example, exact-zero scores for irrelevant edges are certainly desirable, yet it is known that it is hard to deal with exact zeros in common gradient-based architectures. 
A similar argument can be made for non-local aggregations, which are typically incorporated in the detector to improve its expressivity, meaning that its removal is expected to be detrimental. This clearly questions the design of current modular GNNs, along with an inherent trade-off between expressivity and faithfulness, as already outlined in the main text.
In summary, the aim of this empirical investigation was to highlight that some common and often unnoticed design choices play a relevant role in faithfulness, hoping that our results can guide researchers in developing more trustworthy GNNs while pointing out possible dead-ends.  We plan to investigate a de-novo faithful-by-design architecture implementing all desiderata in future work.}

\correction{}{Given the considerations above, we describe in the following how the four desiderata are actually implemented:}

\begin{itemize}[leftmargin=1.25em]

    \item \textbf{Hard Scores} (\textbf{HS}) 
        To enforce the generation of binary 0-1 explanation masks by the detector \DET, we apply a technique similar to the Straight-Through (ST) trick used in the discrete Gumbel-Softmax Estimator \cite{jang2017categorical}. Specifically, during the forward pass, we utilize the binary version of the mask, while in the backward pass, we use the continuous version.

    \correction{}{\item \textbf{Content Features} (\textbf{CF}) To ensure the classifier uses only raw input features, we simply replace the feature matrix with that of the input data before feeding it to the classifier \CLF.}

    \item \textbf{Explanation Readout} (\textbf{ER})
        The Explanation Readout strategy computes the graph embedding by applying a global aggregator \correction{solely to $\RX$}{encouraging the classifier to adhere more to $\RX$}. This involves multiplying the node mask $M$ \correction{}{, obtained by averaging over incident nodes the predicted edge mask,} with the node embedding before performing the final global readout \correction{}{, i.e., $\vh_G = \aggr_G(\{\!\{ M_u\vh_u^L : u \in V \}\!\})$}. Since explanations are soft scores over $\GX$, this approach ensures the model adheres more closely to the soft mask.

    \item \textbf{Local \aggr} (\textbf{LA})
        To enhance the expressivity of Graph Neural Networks (GNNs), some models incorporate virtual nodes \cite{barcelo2020logical,hu2020open,sestak2024vnegnn}. We remove those to mitigate the risk of mixing the information of the explanation with that of its complement, which can create unwanted dependencies between pairs of nodes in the graph.

\end{itemize}

\paragraph{Changes with respect to the original codebase} Here we describe two minor changes we did to the original codebase.\\
\textit{Stable TopK \& Permutation-Invariant Metrics.} We found the original implementation of the topK operator to exhibit instabilities when used on GPU, in particular in the presence of equal scores for which alternatively the first or the last elements of the tensor are returned.
This results in order-dependent metric values, either over -or under-estimated according to the order of ground truth edges in the graph\footnote{In synthetic datasets, ground truth edges are typically the last elements as they are attached to an already generated base-graph.}.

To avoid this bias and to have more predictable behavior, we switched to a stable implementation of the topK operator which always selects elements with the same score as they appear in the input tensor, and we randomly permute nodes and edges in each graph at loading time.

\textit{Undirected Explanation Scores.} Accordingly to the original version of the codebase introduced in Training and Reproducibility paragraph, \LECI, and \GSAT are averaging the edge attention scores among directions for each undirected edge via \texttt{torch\_sparse.transpose}. However, we found that, even by sticking to the original package versioning, the way \texttt{torch\_sparse.transpose} is used is not weighting edge scores as expected, but rather is acting as an identity mapping. To fix this bug and to produce undirected edge scores without changing the edge attention mechanism, we average the edge scores via \texttt{torch\_geometric.to\_undirected}. More details are available in our codebase. 

\cref{tab:directed_vs_undirected_acc} compares performance scores for directed vs undirected explanation scores showing comparable performance aggregated over all datasets and models.

\begin{table}[htbp]
\centering
\scriptsize
\caption{Model performance scores for directed vs undirected explanation scores. Models labeled with '(D)' use directed edge scores.}
\label{tab:directed_vs_undirected_acc}
\begin{tabular}{lcccccc}
\toprule
Dataset / Model & \LECI & \CIGA & \GSAT & \LECI (D) & \CIGA (D) & \GSAT (D) \\
\midrule
\MotifBasis    & \nentry{72}{06} & \nentry{42}{02} & \nentry{57}{03} & \nentry{82}{05} &\nentry{50}{05} & \nentry{52}{04}\\
\MotifTwoBasis & \nentry{85}{07} & \nentry{46}{10} & \nentry{75}{06} & \nentry{81}{06} &\nentry{40}{03} & \nentry{77}{05}\\
\MotifSize     & \nentry{41}{06} & \nentry{43}{05} & \nentry{51}{03} & \nentry{54}{06} &\nentry{47}{02} & \nentry{52}{04}\\
\SSTTwo        & \nentry{83}{01} & \nentry{76}{06} & \nentry{79}{04} & \nentry{83}{01} &\nentry{77}{04} & \nentry{81}{02}\\
\LBAPcore      & \nentry{71}{00} & \nentry{69}{01} & \nentry{70}{00} & \nentry{72}{00} &\nentry{70}{01} & \nentry{71}{00}\\
\CMNIST        & \nentry{26}{10} & \nentry{23}{03} & \nentry{25}{04} & \nentry{28}{17} &\nentry{21}{03} & \nentry{38}{04}\\
\bottomrule
\end{tabular}
\end{table}

\section{Further Discussion}

\correction{}{\subsection{Are OOD perturbations always a bad thing?}}

\correction{}{A recent trend in the XAI literature is focusing on devising perturbation methods guaranteeing that the intervened graph is not OOD, in the sense that the resulting graph is not \textit{too much different} from training graphs, deeming this as the \textit{OOD problem of XAI evaluation} \cite{amara2023ginxeval, fang2023evaluating, qiu2021resistingoutofdistributiondataproblem, hase2021oodxaiproblem}.
In this complement we question the above terminology, arguing that in some circumstances evaluating, e.g. faithfulness, with OOD samples is useful.
An immediate example comes from \cref{thm:faith_ood_upper_bound}, which prescribes that if explanations are both plausible and faithful in ID and OOD data then the model will achieve a consistent behavior across the two data splits.
Therefore, computing OOD faithfulness requires either to have access to OOD data or to a convenient way to simulate it.
Applying controlled interventions that can bring graphs OOD is therefore useful for this evaluation.
Another example is in faithfulness certification: as discussed in \cref{prop:metrics-are-not-equivalent} the faithfulness certificate one can compute on a certain distribution of explanation complements $\PR$ might not hold for a different distribution $\PR'$. 
Being able to simulate different enough complements can empower stakeholders with certificates that, in practice, can give wider guarantees.
A similar argument is made in \citet{dasgupta2022framework}, where they argue that "\textit{It is not advisable to deploy explainers in real-life settings without verifying faithfulness on the target domain}"
}

\correction{}{
In general terms, OOD does not necessarily mean adversarial, and an OOD sample can be perfectly plausible and as likely to appear in the real world as a training sample, just it happens not to be in the training set.
Therefore, in faithfulness evaluation, one should be more concerned in not generating counterfactual graphs, which are legit to hack the evaluation, and that are naturally present in the training set (e.g. the causal feature of other classes).
}

\correction{}{\subsection{Does explanation invariance entail invariant models?}}

\correction{}{To build intuition, take a \DIGNN that, for some input, outputs a perfectly invariant explanation $R_A$. Now, if the explanation is not strictly sufficient, by definition there exists an OOD modification to the complement of $R_A$ (which is domain-dependent) that alters the predicted class distribution. Thus the model as a whole cannot be domain-invariant.}

\correction{}{This naturally fits \cref{thm:faith_ood_upper_bound}. If the detector outputs perfectly invariant explanations $\lambda_{topo}$ and $\lambda_{feat}$ at the RHS of \cref{eq:faith_ood_upper_bound} are zero, yet the sufficiency term $\lambda_{suff}$ can still be larger than zero, meaning that the model can fail invariance (LHS > 0, e.g., it might fit ID data properly but OOD data poorly).}

\begin{figure}
    \centering
    \subfigure{\includegraphics[width=0.45\textwidth]{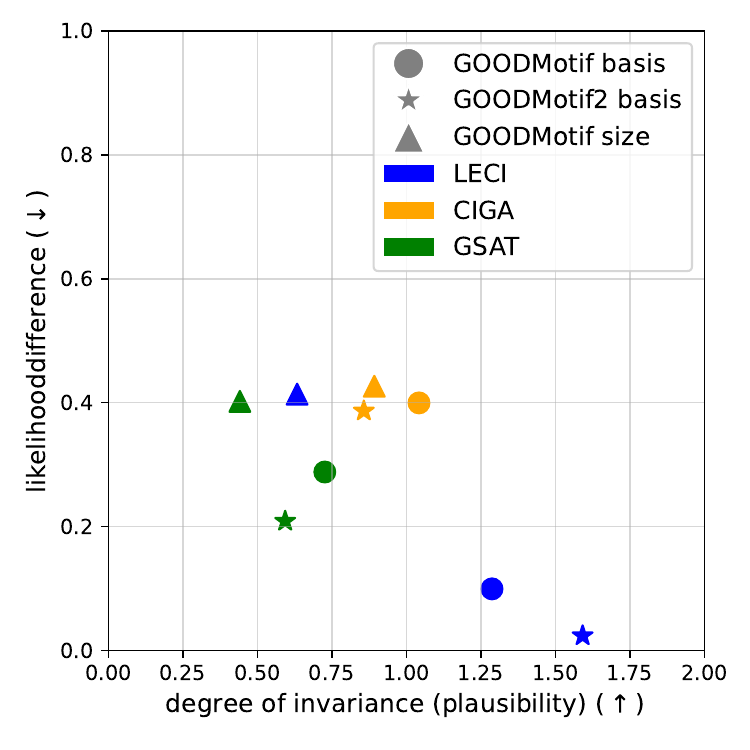}}
    \subfigure{\includegraphics[width=0.45\textwidth]{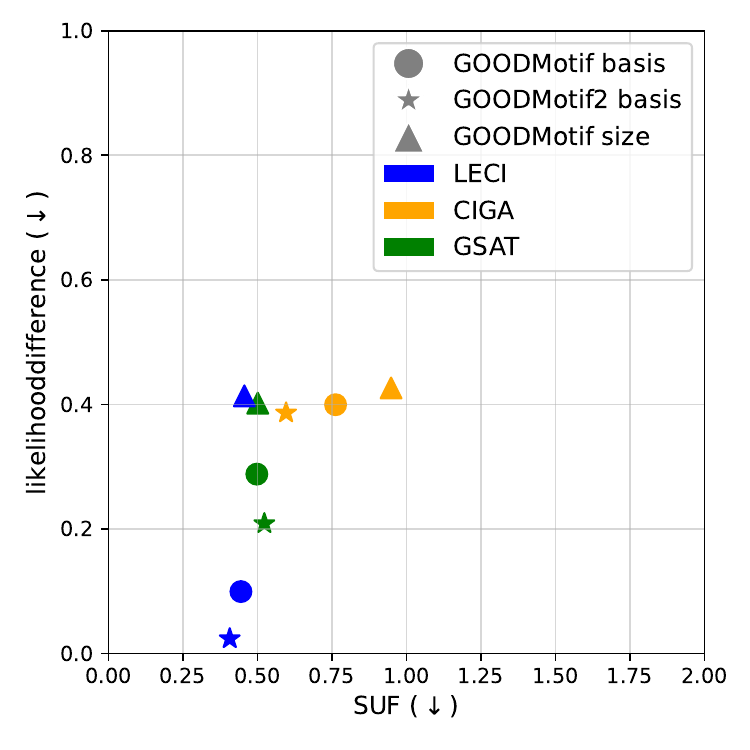}}
    
    \caption{\correction{}{The plots shows the LHS (difference in ID and OOD likelihood) and RHS (invariance/plausibility and sufficiency) of \cref{thm:faith_ood_upper_bound} for nine \DIGNNs used in our experiments on three data sets.
    The x-axis shows that \CIGA (\textcolor{BurntOrange}{orange}) tends to be more domain-invariant (plausible) than \GSAT (\textcolor{ForestGreen}{green}) despite having worse sufficiency.
    The y-axis shows that these models tend to fit ID and OOD data very differently, especially \CIGA despite its explanations being more invariant (plausible) than \GSAT’s.
    The only model achieving considerably low likelihood difference is \LECI (\textcolor{blue}{blue}) for \MotifTwoBasis and \MotifBasis, which scores the best both in explanation invariance and \SUF.
    $\downarrow$ ($\uparrow$) stands for the lower (the higher) the better.}}
    \label{fig:lhs_vs_rhs}
\end{figure}

\correction{}{We illustrate the relationship between the RHS and LHS in \cref{fig:lhs_vs_rhs}, which shows that despite \CIGA having more invariant explanations than \GSAT, it has worse sufficiency and they tend to fit ID and OOD with a comparable shift. The plots report the LHS (difference in ID and OOD fit) and RHS (invariance/plausibility and sufficiency) of nine \DIGNNs used in our experiments on two data sets. The x-axis shows that \CIGA (yellow) tends to be more domain-invariant (plausible) than \GSAT (green) despite having worse sufficiency. The y-axis shows that these models tend to fit ID and OOD data very differently, especially \CIGA despite its explanations being more invariant (plausible) than \GSAT~’s. }

\correction{}{As expected, \textbf{explanation invariance does not entail models are domain invariant}.}

\correction{}{\subsection{Can we achieve strict faithfulness in modular GNNs?}}
\label{app:strict_faith_modular}

\correction{}{In \cref{prop:sse-are-trivial} we provide a theoretical argument regarding the impossibility of injective regular GNNs to achieve strict faithfulness.
\textit{Can we provide a similar result for modular GNNs as well?}
We can answer this question with a constructive argument:
Assume a perfectly stable detector \DET, \ie it always predicts the same $\RX$ for all $\GX' \sim \PR$.
Assume also an injective classifier \CLF such that it implements \textbf{HS}, \textbf{CF}, and \textbf{ER}. 
Then, intuitively, the classifier uses no information outside of $\RX$, as the features it is aggregating solely belong to the explanation itself.
Then, as $\RX$ is not changing after perturbations of the complement by the stability assumption, every deterministic GNN will achieve strict sufficiency.
Then, by injectivity of \CLF,  every modification to $\RX$ will change the model output, meaning $\RX$ is also strictly necessary.}

\correction{}{This nice result, however, hinders a number of implementation challenges, some of which are already discussed in \cref{app:impl_details}. In addition, also the possibility of learning truly injective GNN classifiers has recently been questioned \cite{jaeger2023learning}, making this result more of a conceptual blueprint than an actual model design.}

\begin{figure}[]
    \centering
    \subfigure{\includegraphics[width=0.7\textwidth]{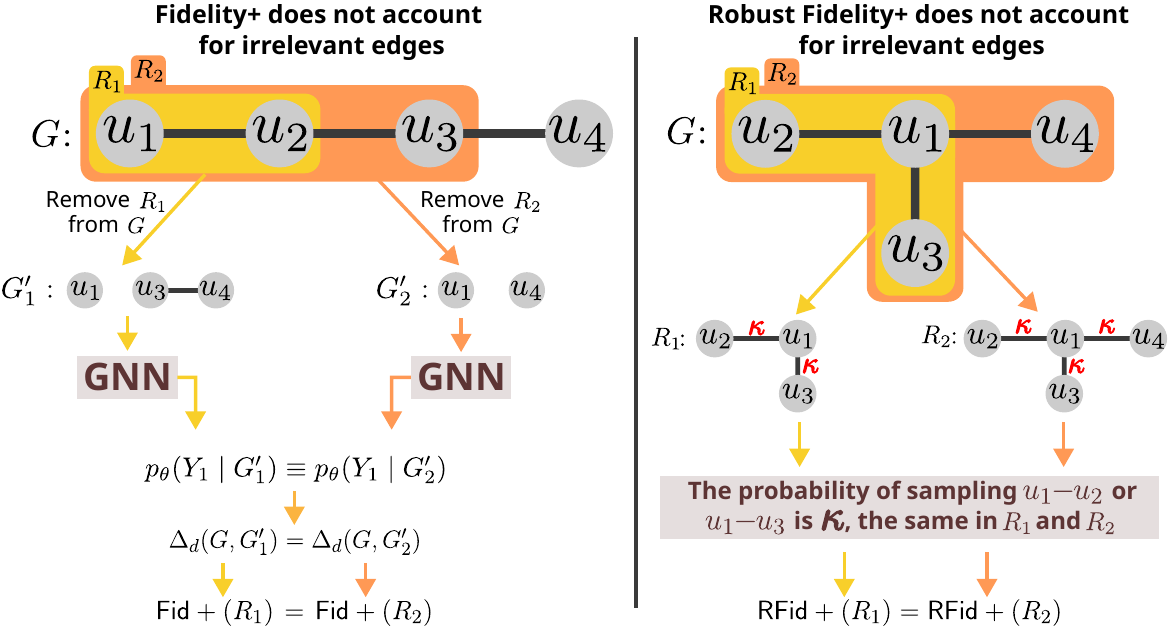}}
    \caption{\correction{}{Illustrative example of why \FIDP and \RFIDP are insensitive to the number of irrelevant edges. Details about the two examples are provided in \cref{sec:estimators_not_equally_good}.}}
    \label{fig:illustration}
\end{figure}

\correction{}{\subsection{Pitfalls of \FIDP and \RFIDP}} \label{app:example_illustration}
\correction{}{In \cref{fig:illustration} shows an illustration of the two examples presented in \cref{sec:estimators_not_equally_good} showing the insensibility of \FIDP and \RFIDP to irrelevant edges.}


\section{Further Experiments}
\label{app:further_experiments}

\correction{}{\subsection{How unstable can \DIGNNs get?}}

\begin{figure}[H]
    \centering
    \includegraphics[width=0.45\textwidth]{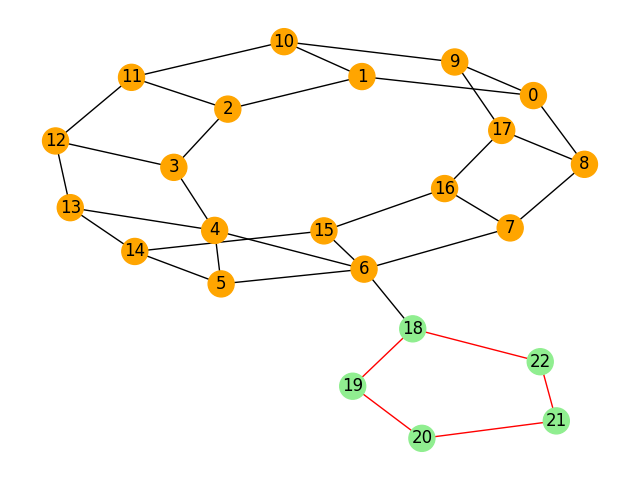}
    \includegraphics[width=0.45\textwidth]{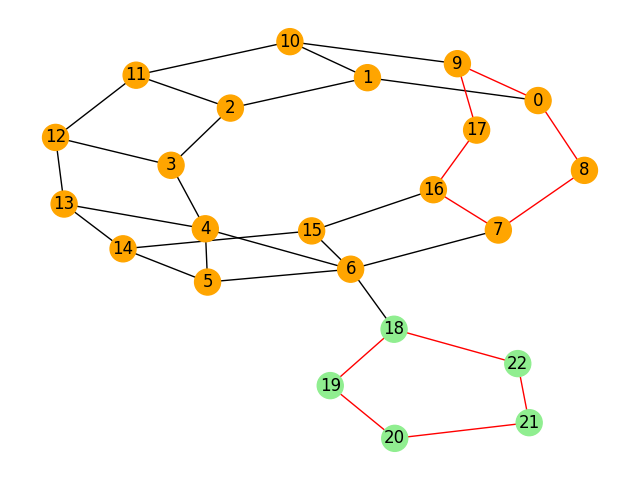}
    \caption{\correction{}{A selected example from the validation set of \MotifTwoBasis for \LECI (seed 1). The left figure shows the predicted explanation on the clean sample, while the right figure shows the explanation after removing edge $(17,8)$. Explanation scores are thresholded at $0.8$ and shown in \textcolor{red}{red}, while the ground truth explanation is highlighted with \textcolor{green}{green} nodes.  Therefore, \textbf{the explanation is not stable to changes to its complement}.}}
    \label{fig:example_stability}
\end{figure}

\correction{}{In \cref{fig:example_stability} we report an example from \MotifTwoBasis for \LECI showing how a single edge removal outside of the explanation can greatly change the explanation output by the detector. As discussed in the main text, the instability of explanations is a factor of model unfaithfulness.}

\subsection{Failure case}
\label{app:failure_cases}

\begin{table}[htbp]
\centering
\scriptsize
\caption{Model performance scores for \SSTTwo on ID validation set and OOD test set.}
\label{table:all_acc}
\begin{tabular}{lcccccc}
\toprule
\multirow{2}{*}{Dataset / Model}
 & \multicolumn{2}{c}{\GIN} & \multicolumn{2}{c}{\LECI} & \multicolumn{2}{c}{\GSAT} \\
\cmidrule(lr){2-3} \cmidrule(lr){4-5} \cmidrule(lr){6-7} 
 & ID val. & OOD test & ID val. & OOD test & ID val. & OOD test \\
\midrule
\SSTTwo         & \nentry{91}{00} & \nentry{80}{01} & \nentry{92}{00} & \nentry{83}{01}   & \nentry{91}{00}& \nentry{79}{04}\\
\bottomrule
\end{tabular}
\end{table}

\begin{figure}
    \centering
    \subfigure{\includegraphics[width=0.8\textwidth]{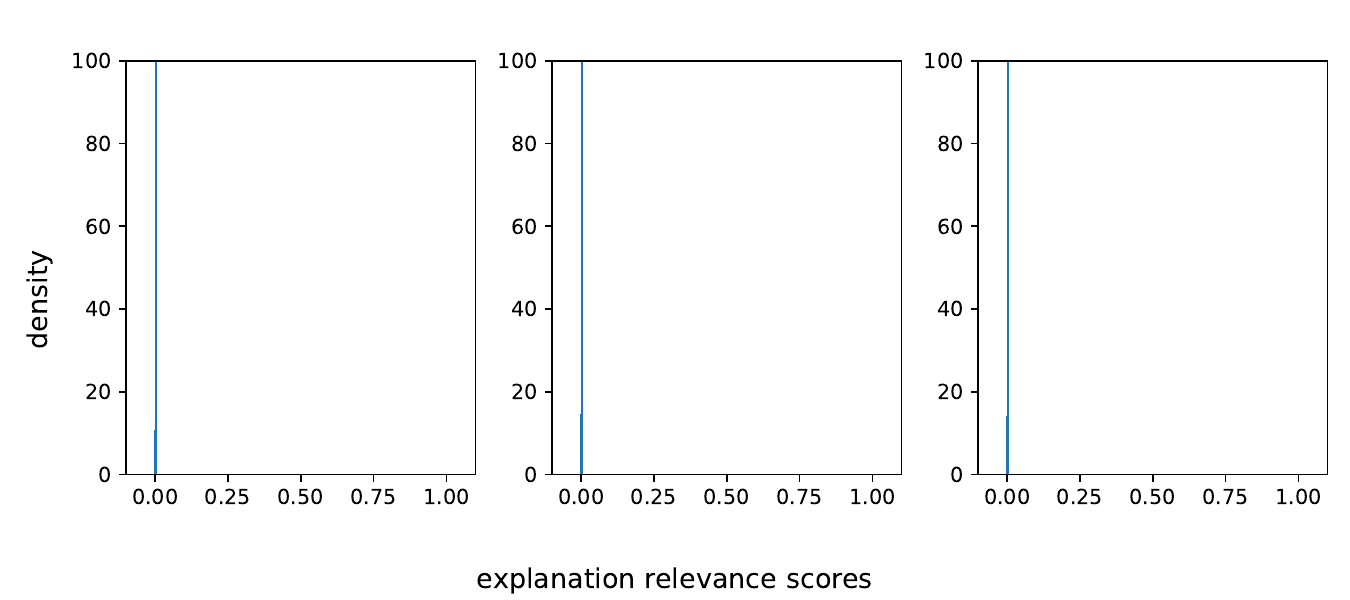}}
    \subfigure{\includegraphics[width=0.8\textwidth]{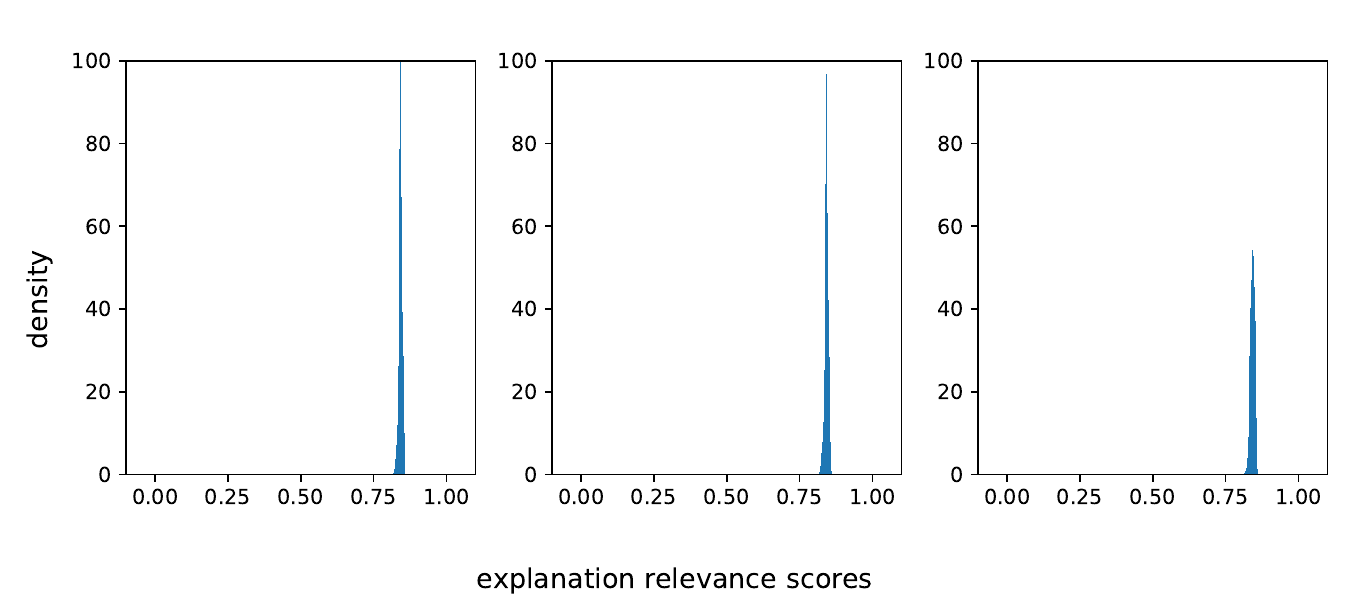}}
    \caption{\textbf{Histograms of explanation relevance scores for \LECI (top), and \GSAT (bottom) on \SSTTwo} (seed $1$). 
    Both models failed in identifying a sparse input subgraph, assigning constant scores (or very close thereof) to every edge.
    %
    }
    \label{fig:expl_histograms}
\end{figure}

As discussed in \cref{sec:perfect_faith_modular}, \SSTTwo is a particularly challenging dataset for improving faithfulness.
We claim this may come from the task benefiting from global information being used, which clashes with the goal of modular architectures that seek sparse and local discriminative subgraphs.
Consider, in fact, the results reported in \cref{table:all_acc} where even a simple monolithic GNN baseline trained without any invariance-aware strategies exhibits comparable performance compared to more advanced and complex \DIGNNs.
To inspect where \DIGNNs fall short, we plot in \cref{fig:expl_histograms} the distribution of edge scores predicted by the detector of \LECI and \GSAT, where it is clear that they failed in identifying any sparse invariant subgraph, 
therefore falling back to a regular-GNN-like behavior. 
%
%
This is \correction{especially}{arguably} expected for \SSTTwo, which is a graph-adapted sentiment classification dataset where node features are contextual embeddings extracted from a BERT-like Transformer \citep{yuan2022explainability} -- which produces intrinsically globally-correlated node representations by the nature of its self-attention layers \citep{vaswani2017attention}. Indeed, BERT-like Transformers are the models achieving state-of-the-art performance in this task \citep{yuan2022explainability}. 


\begin{figure}
        \includegraphics[width=0.45\textwidth]{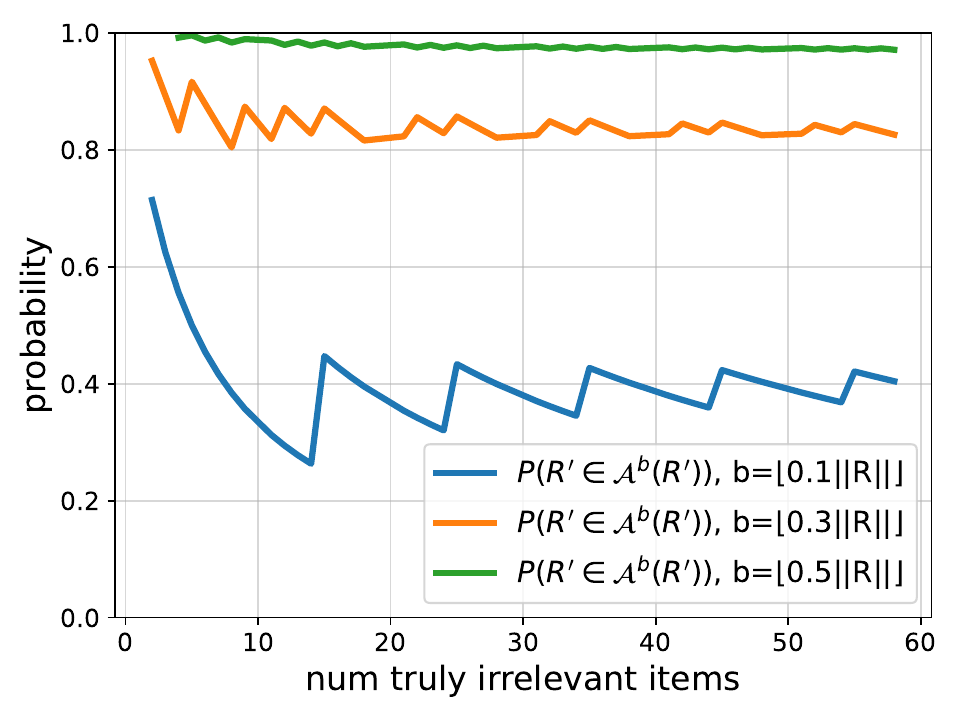}
    \caption{
    \textbf{Probability of deleting at least one truly relevant edge is independent of the number of irrelevant edges if 
    the deletion budget depends on the explanation size.}
    Given an explanation $R$ with $r$ truly relevant edges ($r=5$), and a budget \budget proportional to the size of the explanation, the plot shows $P(R' \in \calA^\budget(R'))$ where $R' \sim \PC^\budget(G)$, for a growing number of irrelevant edges in $R$. The plot shows that the probability is approximately constant, i.e., it does not depend on the number of irrelevant edges. The segments with decreasing behaviour (especially visible for a $10\%$ budget, the \textcolor{blue}{blue} curve) correspond to areas where the budget is indeed constant, and thus not proportional to the explanation size. For instance, between 1 and 14 irrelevant edges, a budget of $10\%$ corresponds to deleting one edge.
  \label{fig:budget-and-probability}}
\end{figure}

\subsection{Sensitivity of \RFIDP and \NEC to the explanation size}
\label{app:further_experiments_nec}

In~\cref{fig:budget-and-probability}, we numerically simulate the probability of deleting at least one truly relevant edge from an explanation with a growing number of irrelevant ones. The figure shows that the probability is basically insensitive to the number of irrelevant edges.

\begin{figure}[]
    \centering
    \subfigure{\includegraphics[width=0.8\textwidth]{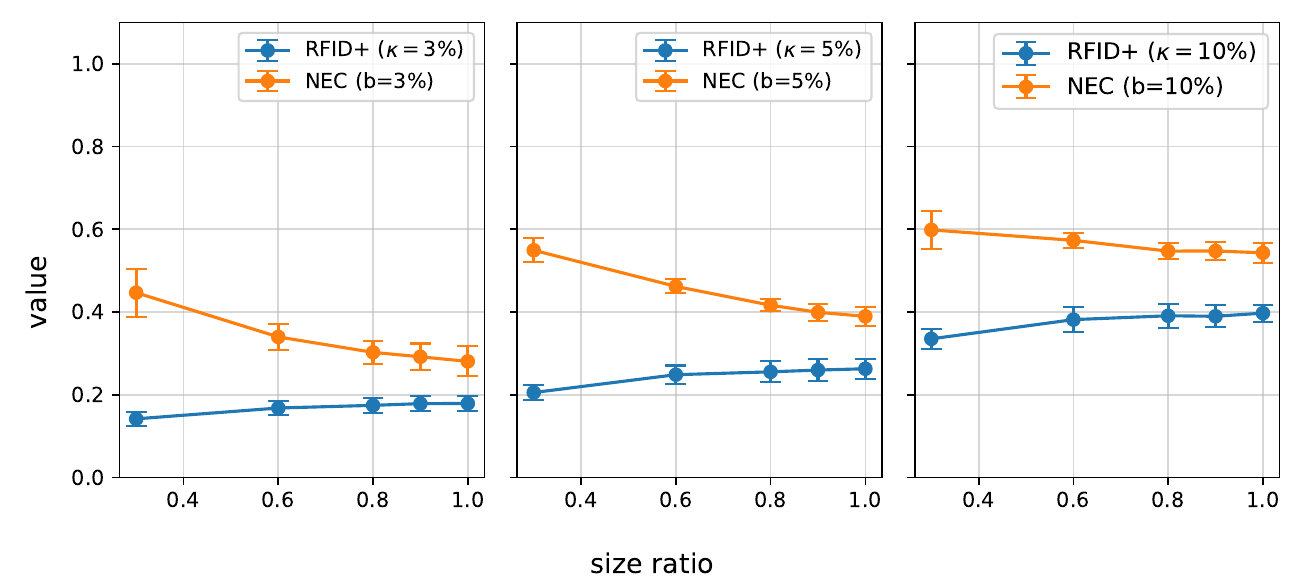}}
    \subfigure{\includegraphics[width=0.8\textwidth]{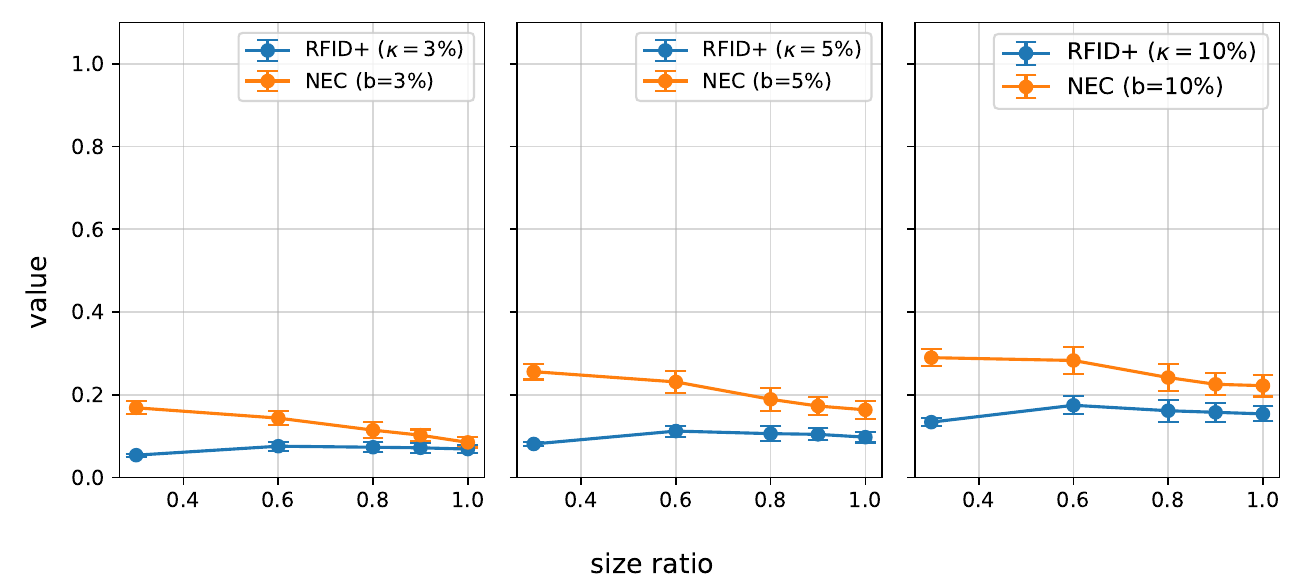}}
    \caption{\textbf{\RFIDP is insensitive to irrelevant edges}. \RFIDP and \NEC with $\PC^\budget$ are computed for \LECI and averaged across 5 seeds on \MotifTwoBasis (top) and \LBAPcore (bottom) for different explanation sizes (x-axis) and for different metric hyper-parameter $\kappa, \budget \in \{ 3\%, 5\%, 10\% \}$. \RFIDP assigns similar or even higher scores to larger explanations, while \NEC tends to penalize larger explanations.}
    \label{fig:sampling_strategies_nec}
\end{figure}

In \cref{fig:budget-and-insensitivity} (Right) we provided an experiments showing that for \GSAT, \RFIDP is in fact insensible to the number of irrelevant edges in the explanation, while \NEC with a suitable \PC is not. In \cref{fig:sampling_strategies_nec} and \cref{fig:sampling_strategies_nec2} we further show that this behavior is in fact general, and occurs across different models and datasets. Specifically, for the same setting delineated in \cref{sec:estimators_not_equally_good}, we report the results also for the \LECI \cite{gui2023joint} model over \LBAPcore \cite{gui2022good} molecular benchmark. 
\begin{figure}[]
    \centering
    \subfigure{\includegraphics[width=0.8\textwidth]{figures/sampling_types/dev_nec_sampling_GSAT_GOODMotif2_basis_test.pdf}}
    \subfigure{\includegraphics[width=0.8\textwidth]{figures/sampling_types/dev_nec_sampling_GSAT_LBAPcore_assay_test.pdf}}
    \caption{\textbf{\RFIDP is insensitive to irrelevant edges}. \RFIDP and \NEC with $\PC^\budget$ are computed for \GSAT and averaged across 5 seeds on \MotifTwoBasis (top) and \LBAPcore (bottom) for different explanation sizes (x-axis) and for different metric hyper-parameter $\kappa, \budget \in \{ 3\%, 5\%, 10\% \}$. \RFIDP assigns similar or even higher scores to larger explanations, while \NEC tends to penalize larger explanations.}
    \label{fig:sampling_strategies_nec2}
\end{figure}


\end{document}